\documentclass[10pt,twocolumn,letterpaper]{article}

\usepackage[pagenumbers]{cvpr} 

\definecolor{cvprblue}{rgb}{0.21,0.49,0.74}
\usepackage[pagebackref,breaklinks,colorlinks,allcolors=cvprblue]{hyperref}
\usepackage{tikz} 
\usetikzlibrary{arrows.meta}
\newcommand{\ours}{Ours}
\usepackage{multirow}
\def\confName{CVPR}
\def\confYear{2027}
\usepackage{placeins} 
\title{HyMTS: A Multiscale Tensor Factorization-Based Network for Hyperspectral Image Restoration}

\author{%
  Lei Xu\textsuperscript{1} \quad
  Fahad Sohrab \textsuperscript{2} \quad
  Moncef Gabbouj\textsuperscript{1} \quad
  Mehmet Yamac \textsuperscript{1} \quad
  \\[0.5em]
  \textsuperscript{1} Tampere University, Finland \quad 
  \textsuperscript{2} Tampere University of Applied Sciences, Finland \\
  \texttt{\{lei.xu, fahad.sohrab, mehmet.yamac, moncef.gabbouj\}@tuni.fi} \quad
}
\begin{document}
\maketitle
\begin{abstract}
Hyperspectral image (HSI) restoration has been explored using deep learning
techniques recently. Because hyperspectral images have strong spectral and
spatial redundancy, the underlying clean signal is usually represented in a
low-dimensional spectral subspace and locally low-rank spatial structures.
Existing methods primarily explore dense attention or feature mixing across
spectral bands. In this work, we propose HyMTS, a hyperspectral restoration network with a backbone built on the Multi-scale Tensorial Summation (MTS) factorization operator. HyMTS provides a natural architectural mechanism for exploiting a low-rank prior, embedded directly in the network operator rather than imposed via an auxiliary loss or a constraint on the reconstructed output. Specifically, HyMTS stacks a unified block that couples factorized layers with input-adaptive layers, capturing both local and global dependencies across the spectral and spatial dimensions. Extensive experiments without auxiliary data on standard benchmarks for Gaussian and real-noise denoising and single-image super-resolution show that HyMTS efficiently achieves superior or comparable performance to various state-of-the-art approaches. 
\end{abstract}    
\section{Introduction}
\label{sec:intro}
Hyperspectral images (HSI) scenes usually contain a wide and continuous range of the electromagnetic spectrum with multiple spectral bands~\cite{leixuhsi}. Hence, HSI enable fine-grained information for various applications, such as mineral detection~\cite{mineralHSI}, medical analysis~\cite{ali2025surgicalhsi} and agricultural inspections~\cite{ram2024systematic}. Because of redundancy in the spectral and spatial dimensions, it is essential to use strong structural priors~\cite{leixuhsi, Bajwa2004Hyperspectral,maskedHSI} for various HSI analyses.
\begin{figure}[ht]
  \centering
  \includegraphics[width=1.0\linewidth]{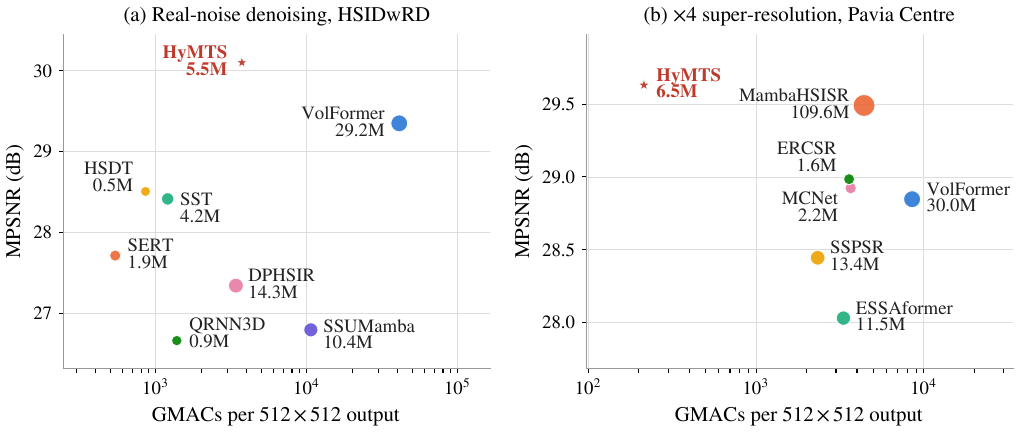}
  \caption{PSNR versus GMACS for a $512\times512$ cube.}
  \label{fig:eff}
\end{figure}
Therefore, a key challenge in successfully restoring HSI is balancing spatial detail recovery with spectral fidelity~\cite{kolyszko2026esspn,yu2025volformer}. 
Existing HSI restoration methods can be mainly categorized into traditional-based methods, CNN-based methods, transformer-based methods, diffusion-based methods, and Mamba-based methods. Traditional methods learn various priors~\cite{vrltd,nlmg,shlg}, such as spectral subspace, tensor factorization, or sparsity. They then use these learned priors to exploit the spatial and spectral properties of the hyperspectral cube. CNNs-based methods utilize various convolutional neural networks (CNNs) structures to capture spatial-spectral features from paired input data~\cite{torun2024smcnn,3dqrnn,hsisupercnn,Mei2017HyperspectralIS,Hsi-denet}. CNN-based methods have achieved promising results. Nevertheless, CNNs have clear disadvantages: their receptive field is inherently limited, and full 3D operations are computationally expensive~\cite {3dqrnn}. As a long-range dependency approach~\cite{transformer}, transformer-based methods have recently been widely developed for various HSI restoration tasks. The self-attention structures capture long-range spatial and spectral dependencies through adaptive interactions~\cite{essaformer,yu2025volformer,sert}. However, transformer-based methods require specific complexity-based attention mechanisms~\cite{Wu_2025_ICCV}. Mamba~\cite{gu2024mamba} is a linear-time sequence model that uses selective state-spaces to learn long-range dependencies, making it attractive for HSI restoration. The performance of Mamba-related methods usually depends on how efficiently the scanning order is designed for sequence construction~\cite{Chen2026USPMambaUS,mambahsisr,lamamba}. In addition, diffusion-based models~\cite{hir-diff,Jiang_2026_CVPR,Miao2023DDS2MSD} have been used for ill-posed HSI restoration due to their strong generative priors and robustness to noise. However, diffusion-based models are challenging not only to preserve spectral fidelity but also to train due to their computational expense from iterative sampling~\cite{hir-diff,Jiang_2026_CVPR}. 

To address these limitations, we propose HyMTS, a Multi-scale Tensorial Summation (MTS) factorization operation-based network for HSI restoration. Yama\c{c} et al.~\cite{yamacc2025multiscale,edgemts} introduced MTS as a backbone layer that sums Tucker-type mode products~\cite{tucker1966,kolda2009,efficienttd} over windows at multiple spatial scales, providing large receptive fields and cross-dimensional correlations at low parameter cost~\cite{yamac2026mtscsnet}. The proposed HyMTS uses compact, shared multilinear projections of MTS to local spectral-spatial correlations in HSI scenes. HyMTS repeats one block that combines three kinds of layers. Static layers are MTS maps with fixed weights: two multi-scale tensorial-summation layers capture local spectral–spatial structure, and a strip-window layer applies the same operator to (channel × row) and (channel × column) planes so that channel mixing depends on spatial position. Gated layers modulate fixed weights by the input: a bidirectional recurrent scan links each pixel to its whole row and column, and a high-pass gated feed-forward network adds zero-mean detail. Adaptive layers compute their weights from the input: shifted-window self-attention selects neighbours within and across windows, and pooled spectral attention reweights channels per cube. The same backbone serves denoising and super-resolution; only the output head changes. The contributions of this work are summarized as follows:
\begin{itemize}
    \item We propose a novel HyMTS based on a compact MTS architecture for HSI denoising and super-resolution. Its shared multilinear projections couple spatial and spectral information with few parameters per receptive field.
    \item We introduce several mechanisms to capture spectral-spatial dependencies globally and locally. With multi-scale tensorial-summation and strip-window coupling, we learn local spectral--spatial coupling. The attention unit, shifted-window self-attention followed by a high-pass gated feed-forward network, exposes cross-window interactions. The pooled spectral attention reweights the feature channels by the content of the whole cube. The bidirectional recurrent scan captures the global spatial context.
    \item Extensive experiments on Gaussian denoising, real-noise denoising, and super-resolution show that HyMTS achieves state-of-the-art accuracy efficiently, especially for HSI super-resolution.
\end{itemize}

\section{Related works}
\label{sec:related}
\subsection{HSI restoration}
HSI restoration, including HSI denoising and HSI super-resolution, aims to restore high-quality hyperspectral images from degraded ones. The HSI denoising task removes noise from corrupted inputs while preserving spatial resolution and spectral fidelity. HSI super-resolution aims to reconstruct high-resolution HSI scenes from their low-resolution inputs. Traditional methods usually frame this task as an optimization problem and investigate underlying HSI denoising formulations. For example, Zhang et al.~\cite{zhang2014lrmr} use low-rank matrix recovery to remove various noises, while~\cite {nlmg} explores global subspace estimation in cooperation with non-local spatial denoising. CNN-based methods have been introduced to tackle HSI denoising and subsequently improve spatial-spectral representation. Wei et al. propose a 3D quasi-recurrent neural network to extract structural spatio-spectral correlation in an HSI scene for HSI denoising~\cite{3dqrnn}. HSI-DeNet~\cite{Hsi-denet} is based on a fully CNN structure with learned multichannel filters in each layer to explicitly model spatio-spectral correlation. Recently, transformer-based methods have achieved promising outcomes. SERT~\cite{sert} uses a spectral-enhanced rectangle transformer to capture long-range dependencies for HSI denoising. SST~\cite{sst} models non-local characteristics of HSI scenes with improved computational efficiency and capability based on a spatial-spectral transformer. ESSAformer~\cite{essaformer} explores an efficient SCC-kernel-based self-attention (ESSA) embedded transformer network for single HSI super-resolution. VolFormer~\cite{yu2025volformer} proposes a unified HSI restoration backbone that jointly models spatial and spectral interactions using volumetric self-attention. In addition, Mamba-based methods have emerged as an alternative for modeling long-range dependencies. LaMamba~\cite{lamamba} introduces a Mamba backbone with linear attention for HSI denoising. MambaHSISR~\cite{mambahsisr} is a Mamba HSI super-resolution framework with three components to capture long-range dependencies and spatial-spectral details. Diffusion-based methods have been used for HSI restoration using generative priors. HIR-Diff~\cite{hir-diff} is an unsupervised HSI restoration framework that uses a pre-trained diffusion model working with two low-rank components and a total variation prior. 

\subsection{MTS factorization Network}
Multi-scale Tensorial Summation (MTS)~\cite{yamacc2025multiscale} is an efficient neural network operator for multidimensional data processing that directly performs structured multilinear transformations on input features. Specifically, it partitions features into windows at multiple scales and applies shared Tucker-like mode-wise projections to capture cross-dimensional interactions with fewer parameters than dense transformations. Based on this operator, Xu et al.~\cite{edgemts, Yama2021GeneralizedTS} propose MTS-DR-Net to capture high-frequency information and suppress redundant features for edge detection. MTS-CSNet~\cite{yamac2026mtscsnet} further explores the MTS operator for RGB compressive sensing using tensorial projection for learnable sampling. EMTS-MHGNet~\cite{EMTS-MHGNet} extends the MTS operator with a multi-scale lightweight architecture for efficient real-world image deblurring. These studies inspire us to use the MTS operator for HSI restoration modeling, balancing model efficiency and effectiveness.

\section{Proposed methods}  
\subsection{Notations}
In this work, we denote scalars, vectors, matrices, and third-order tensors (HSI cubes) as $y$, $\mathbf{y}$, $\mathbf{Y}$, and $\mathcal{Y}$, respectively. A hyperspectral cube with $H\times W$ pixels and $C_b$ bands is denoted as: $\mathcal{Y}\in\mathbb{R}^{C_b\times H\times W}$. The $b$-th band of $\mathcal{Y}$ is $\mathbf{Y}_b\in\mathbb{R}^{H\times W}$ and the spectrum at pixel
$(i,j)$ is $\mathbf{y}_{ij}=\mathcal{Y}^{(:,i,j)}\in\mathbb{R}^{C_b}$. The network's intermediate feature cube with $C$ channels is denoted as $\mathcal{F}_{i}\in\mathbb{R}^{C\times H\times W}, i\in{0,...,N}$. $\mathcal{Y}$ is the ground truth, $\hat{\mathcal{Y}}$ is the corresponding degraded HSI cube, and $\Tilde{\mathcal{Y}}$ is the output of the HyMTS network. For super-resolution, $\hat{\mathcal{Y}}$ denotes the low-resolution input of size $H\times W$, and $\mathcal{Y}$ and $\Tilde{\mathcal{Y}}$ has size $sH\times sW$, where $s\in\{4,8\} $ is a scale factor. For denoising, $\mathcal{Y}$, $\Tilde{\mathcal{Y}}$, and  $\hat{\mathcal{Y}}$ all have size $H\times W$
\begin{figure*}[ht]
  \centering
  \includegraphics[width=1.0\linewidth]{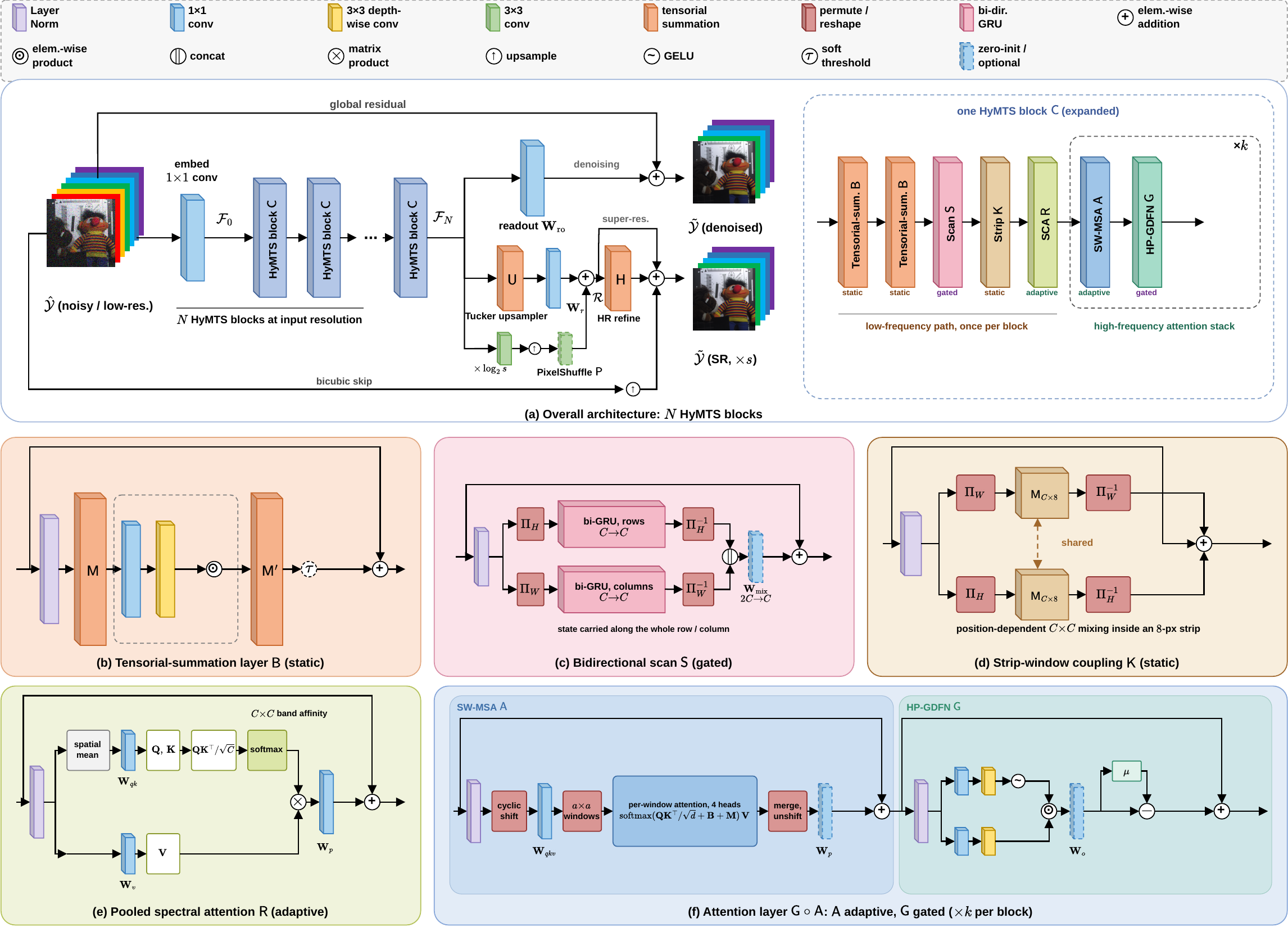}
  \caption{The architecture of the proposed HyMTS.}
  \label{fig:schema}
\end{figure*}
\subsection{Overall Architecture}
The whole HyMTS network applies to a degraded or low-resolution HSI cube $\hat{\mathcal{Y}}
\in\mathbb{R}^{C_b\times H\times W}$ as
\begin{equation}
\Tilde{\mathcal{Y}}=\mathsf{D}\big(\mathsf{C}^{N}(\mathsf{E}(\hat{\mathcal{Y}})),\,\hat{\mathcal{Y}}\big),
\label{eq:overall}
\end{equation}
where $\mathsf{E}$ is a linear embedding, $\mathsf{C}^{N}$ denotes the backbone with $N$ HyMTS blocks. $\mathsf{D}$ denotes the final task head for denoising or for super-resolution. As shown in Fig.~\ref{fig:schema}(a), the input HSI cube $\mathcal{Y}$ is firstly projected from $C_b$ bands to $C$ channels with a $1\times 1$ convolution without bias, as $\mathcal{F}_0=\mathsf{E}(\hat{\mathcal{Y}})$.
\subsection{HyMTS Block and HyMTS Backbone}
A HyMTS block maps its input to a feature cube of the same size: $\mathsf{C}:\mathbb{R}^{C\times H\times W}\to\mathbb{R}^{C\times H\times W}$. It has three distinct parts: static layers, gated layers, and adaptive layers, as shown in Tab.~\ref{tab:components} of the supplementary material. Static layers have fixed weights and no activation; gated layers contain fixed weights modulated element-wise by a nonlinear function of the input; adaptive layers have softmax weights computed from the input. A block applies $\mathsf{B}$, $\mathsf{B}$, $\mathsf{S}$, $\mathsf{K}$, and $\mathsf{R}$ in this order, followed by $k$ attention units $\mathsf{G} \circ \mathsf{A}$ (Eq.~\ref{eq:cycle} reads right to left).: 
\begin{equation}
\mathsf{C}
=\big(\underbrace{\mathsf{G}}_{\text{gated}}\circ\underbrace{\mathsf{A}}_{\text{adaptive}}\big)^{k}
\circ\underbrace{\mathsf{R}}_{\text{adaptive}}
\circ\underbrace{\mathsf{K}}_{\text{static}}
\circ\underbrace{\mathsf{S}}_{\text{gated}}
\circ\underbrace{\mathsf{B}\circ\mathsf{B}}_{\text{static}}.
\label{eq:cycle}
\end{equation}
Each component is a pre-norm residual layer with layer normalization (LN), and the attention unit consists of a pair of residual connections with shifted-window multi-head self-attention (SW-MSA) $\mathsf{A}$ and its feed-forward $\mathsf{G}$. The two tensorial-summation layers $\mathsf{B}$ in Fig.~\ref{fig:schema}(b) aim to extract multi-scale spectral-spatial features over a wider receptive field at the limited parameter cost of a factorized map rather than a dense one~\cite{yamacc2025multiscale}. Bidirectional scan $\mathsf{S}$ in Fig.~\ref{fig:schema}(c) spans the whole image along rows and columns to link every pixel to all pixels of its row and column. Strip-window coupling $\mathsf{K}$ in Fig.~\ref{fig:schema}(d) is designed to model spectral dependency with a position-dependent and band-mixing matrix. 
Pooled spectral attention $\mathsf{R}$ in Fig.~\ref{fig:schema}(e)
targets band selection for the cube adaptively. Attention units $\mathsf{G}\circ\mathsf{A}$ in Fig.~\ref{fig:schema}(f), repeated $k$ times, preserve the high-frequency content.

Then, we use HyMTS blocks sequentially $N$ times for the composition of the backbone as
\begin{equation}
\mathsf{C}^{N}=\mathsf{C}_N\circ\cdots\circ\mathsf{C}_1,\qquad
\mathcal{F}_n=\mathsf{C}_n(\mathcal{F}_{n-1}), n \in \{0,\dots,N\}.
\label{eq:backbone}
\end{equation}      
The backbone takes $\mathcal{F}_0$ with the size $C\times H\times W$ as input and produces a feature cube $\mathcal{F}_N=\mathsf{C}^{N}(\mathcal{F}_0)$ with the same size. We use the same backbone for both tasks. $Z = LN(F)$ is the normalized input of the current layer. The head $\mathsf{D}$ produces the final result $\Tilde{\mathcal{Y}}$ with the $C_b$ spectral bands. For HSI denoising head $\mathsf{D}$, the final output is linearly projected to $C_b$ spectral bands and summed with a global residual
\begin{equation}
\tilde{\mathcal{Y}}=\operatorname{clip}_{[0,1]}\big(\hat{\mathcal{Y}}+\mathbf{W}_{\mathrm{ro}}\,\mathcal{F}_N\big),
\label{eq:dn_head}
\end{equation}
where $\mathbf{W}_{\mathrm{ro}}\in\mathbb{R}^{C_b\times C}$ is a $1\times 1$ convolution with a bias, both zero-initialized. For HSI super-resolution head $\mathsf{D}$, the Tucker upsampler $\mathsf{U}$ and a linear projection $\mathbf{W}_{\mathrm{r}}$ map the backbone output $\mathcal{F}_N$ to an output with size $sH\times sW$, which is then added to $\mathsf{PS}(\mathcal{F}_N)$ from a PixelShuffle for $\mathcal{R}$. The final $\Tilde{\mathcal{Y}}$ is a summation with a bicubic skip of the input $\mathrm{Bic}_s(\hat{\mathcal{Y}})$, $\mathsf{Ref}(\mathcal{R})$ from a high-resolution refinement step and $\mathcal{R}$, defined as:
\begin{equation}
\begin{aligned}
\tilde{\mathcal{Y}}&=\operatorname{clip}_{[0,1]}\big(\mathrm{Bic}_{s}(\hat{\mathcal{Y}})+\mathcal{R}+\mathsf{Ref}(\mathcal{R})\big),\\
\mathcal{R}&=\mathbf{W}_{\mathrm{up}}\,\mathsf{U}(\mathcal{F}_N)+\mathsf{PS}(\mathcal{F}_N), 
\end{aligned}
\label{eq:srhead}
\end{equation}
where $\mathbf{W}_{\mathrm{up}}\in\mathbb{R}^{C_b\times C}$ is a $1\times 1$ convolution and $\mathbf{b}_{\mathrm{r}}\in\mathbb{R}^{C_b}$, both zero-initialized. 
\begin{figure*}[ht]
  \centering
  \includegraphics[width=1.0\linewidth]{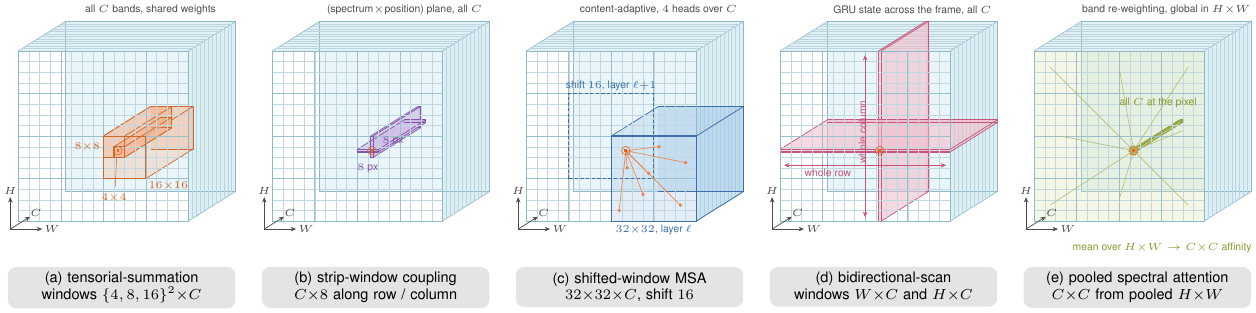}
  \caption{Interaction windows of the operators on one $H{\times}W{\times}C$ cube. (a)~Tensorial summation window: coupled with all $C$ channels by a shared multilinear map.
(b)~Strip-window: an $8$-pixel strip along the row and with the same weights along the column, coupling all $C$ features. (c)~Shifted-window self-attention: a
$32{\times}32$ window, shifted by $16$ pixels in the next layer (dashed).
(d)~Recurrent scan: the only full-frame operator.
(e)~Pooled spectral attention: a $C{\times}C$ band affinity from the spatially averaged
cube, applied at every pixel.}
  \label{fig:windows}
\end{figure*}
\subsection{Multi-scale Tensorial-Summation Layers}
Both the tensorial-summation layers and the SW-MSA operate on non-overlapping windows as shown in Fig.~\ref{fig:windows}. $\mathsf{P}_w$ partitions a feature cube $\mathcal{X}\in\mathbb{R}^{C\times H\times W}$ into non-overlapping $w\times w$ spatial windows as $\mathcal{W}_{(u,v)}\in\mathbb{R}^{C\times w\times w}$ and $\mathsf{P}_w^{-1}$ reassembles windows back to $\mathcal{X}$. We further define 
\begin{equation}
\mathcal{W}_{(u,v)}^{(c,p,q)}=\mathcal{X}^{(c,\;uw+p,\;vw+q)},\qquad
p,q\in\{1,\dots,w\},
\label{eq:part}
\end{equation}
where window is indexed by $u\in\{0,\dots,H/w-1\}$, $v\in\{0,\dots,W/w-1\}$, and $(p,q)$ is the position inside the window. According to~\cite{yamacc2025multiscale}, the tensorial-summation layer at scale $w$ is defined:
\begin{equation}
\mathsf{T}_w(\mathcal{W})=\sum_{t=1}^{T}
\mathcal{W}\times_1 \mathbf{A}^{(w)}_t\times_2 \mathbf{B}^{(w)}_t\times_3 \mathbf{D}^{(w)}_t,
\label{eq:ts}
\end{equation}
where $\mathbf{A}^{(w)}_t\in\mathbb{R}^{C\times C}$ and $\mathbf{B}^{(w)}_t,\mathbf{D}^{(w)}_t\in\mathbb{R}^{w\times w}$. $\mathcal{W}$ denotes a window. $T$ is the number of summed separable terms. The multi-scale operator sums the scales as in Fig.~\ref{fig:windows}(a),
\begin{equation}
\mathsf{M}(\mathcal{X})=\sum_{w\in\Omega}\mathsf{P}_w^{-1}\,
\mathsf{T}_w\big(\mathsf{P}_w \mathcal{X}\big),
\label{eq:ms}
\end{equation}
with $\Omega=\{4,8,16\}$ for denoising and $\{4,8,16,32\}$ for $\times 4$ super-resolution ($\{4,8,16\}$ at $\times 8$, where the low-resolution training patch is $16$ pixels wide). 

Different from~\cite{yamacc2025multiscale}, we redefine the tensorial-summation layers in this work as
\begin{equation}
\mathsf{B}(\mathcal{F})=\mathcal{F}+\mathrm{soft}_{\boldsymbol{\tau}}\Big(\mathsf{M}'\big(g\big(\mathsf{M}(\mathrm{LN}(\mathcal{F}))\big)\big)\Big),
\label{eq:block}
\end{equation}
where the gate $g$ keeps its definition from~\cite{yamacc2025multiscale}. In particular, a per-channel soft threshold sparsifies the correction of each tensorial-summation layer for the super-resolution task, acting as a shrinkage operator
\begin{equation}
\boldsymbol{\tau}=\boldsymbol{\gamma}\odot\operatorname{softplus}(\boldsymbol{\rho}),\qquad
\boldsymbol{\gamma},\boldsymbol{\rho}\in\mathbb{R}^{C},\ \boldsymbol{\gamma}\leftarrow\mathbf{0},
\label{eq:shrink}
\end{equation}
where $\boldsymbol{\gamma}$ is a learned per-channel gain, initialized to zero, and $\boldsymbol{\rho}$ is a learned per-channel parameter with a softplus magnitude threshold.
\begin{table*}[t]
\centering
\caption{Gaussian denoising on ICVL (50 scenes, center $512\times512$ crops,
full-frame inference). All methods are trained by us with the authors'
recipes and read at the last iterate. Blind: $\sigma\sim\mathcal{U}[30,70]$.
GMACs: one $512\times512$ cube (31 bands).}
\label{tab:icvl}
\footnotesize
\setlength{\tabcolsep}{3pt}
\resizebox{\textwidth}{!}{%
\begin{tabular}{@{}l r r ccc ccc ccc ccc ccc@{}}
\toprule
& & & \multicolumn{3}{c}{$\sigma=10$} & \multicolumn{3}{c}{$\sigma=30$} & \multicolumn{3}{c}{$\sigma=50$} & \multicolumn{3}{c}{$\sigma=70$} & \multicolumn{3}{c}{Blind} \\
\cmidrule(lr){4-6}\cmidrule(lr){7-9}\cmidrule(lr){10-12}\cmidrule(lr){13-15}\cmidrule(lr){16-18}
Method & Params & GMACs & MPSNR$\uparrow$ & MSSIM$\uparrow$ & SAM$\downarrow$ & MPSNR$\uparrow$ & MSSIM$\uparrow$ & SAM$\downarrow$ & MPSNR$\uparrow$ & MSSIM$\uparrow$ & SAM$\downarrow$ & MPSNR$\uparrow$ & MSSIM$\uparrow$ & SAM$\downarrow$ & MPSNR$\uparrow$ & MSSIM$\uparrow$ & SAM$\downarrow$ \\
\midrule
HIR-Diff~\cite{hir-diff}$^{a}$& 391.0M & 57567 & 35.365 & 0.9509 & 8.471 & 33.953 & 0.9252 & 6.740 & 28.864 & 0.7913 & 7.768 & 24.837 & 0.6248 & 8.129 & 29.607 & 0.8034 & 7.815 \\ 
QRNN3D~\cite{3dqrnn} & 0.860M & 1257 & 45.181 & 0.9860 & 1.477 & 40.858 & 0.9573 & 2.410 & 38.001 & 0.9206 & 3.329 & 35.647 & 0.8685 & 4.556 & 38.436 & 0.9230 & 3.272 \\ 
DPHSIR~\cite{lai2022dphsir} & 14.277M & 3096 & 45.386 & 0.9878 & 1.295 & 42.335 & 0.9740 & 1.625 & 40.222 & 0.9603 & 1.979 & 38.360 & 0.9429 & 2.502 & 40.495 & 0.9617 & 1.968 \\ 
VolFormer~\cite{yu2025volformer} & 29.174M & 34390 & 47.456 & 0.9916 & 1.016 & 43.978 & 0.9805 & 1.256 & 41.984 & 0.9706 & 1.452 & 40.584 & 0.9620 & 1.622 & 42.342 & 0.9720 & 1.418 \\ 
SST~\cite{sst} & 4.145M & 1198 & 48.939 & \textbf{0.9930} & 0.857 & 44.881 & 0.9825 & 1.148 & 42.665 & 0.9728 & 1.380 & 41.129 & 0.9640 & 1.581 & 43.008 & 0.9742 & 1.352 \\ 
SERT~\cite{sert} & 1.905M & 538 & 48.829 & 0.9929 & 0.846 & 44.937 & \textbf{0.9828} & 1.114 & 42.731 & \textbf{0.9734} & 1.330 & 41.159 & 0.9649 & 1.520 & 43.046 & \textbf{0.9747} & 1.302 \\ 
SSUMamba~\cite{ssumamba} & 10.372M & 9716 & 49.065 & 0.9929 & \textbf{0.801} & 45.002 & 0.9826 & 1.116 & 42.837 & 0.9732 & 1.355 & 41.269 & 0.9649 & 1.569 & 43.113 & 0.9746 & 1.329 \\ 
\textbf{\ours{}} & 5.450M & 3720 & \textbf{49.209} & 0.9929 & 0.820 & \textbf{45.131} & 0.9827 & \textbf{1.092} & \textbf{42.927} & \textbf{0.9734} & \textbf{1.304} & \textbf{41.326} & \textbf{0.9650} & \textbf{1.499} & \textbf{43.263} & \textbf{0.9747} & \textbf{1.275} \\ 
\bottomrule
\end{tabular}}
\begin{flushleft}\footnotesize
$^{a}$ Training-free; last sample reported; GMACs of its 20 sampling steps (pretrained 391.0M diffusion UNet).
\end{flushleft}
\end{table*}
\begin{table}[t]
\centering
\caption{Real-noise denoising on HSIDwRD (44 training / 15 test scenes,
$512\times512$ full-frame). SSIM is the global per-band formula of~\cite{sert}. GMACs: one $512\times512$ cube (34 bands).}
\label{tab:hsidwrd}
\footnotesize
\setlength{\tabcolsep}{4pt}
\resizebox{\linewidth}{!}{%
\begin{tabular}{@{}l r r c c c@{}}
\toprule
Method & Params & GMACs & MPSNR$\uparrow$ & SSIM$\uparrow$ & SAM$\downarrow$ \\
\midrule
QRNN3D~\cite{3dqrnn} & 0.860M & 1378 & 26.666 & 0.9046 & 3.092 \\ 
SSUMamba~\cite{ssumamba} & 10.372M & 10656 & 26.799 & 0.8831 & 2.539 \\ 
DPHSIR~\cite{lai2022dphsir} & 14.277M & 3396 & 27.345 & 0.9140 & 2.528 \\ 
SERT~\cite{sert} & 1.897M & 539 & 27.716 & 0.9267 & 2.840 \\ 
SST~\cite{sst} & 4.150M & 1200 & 28.416 & 0.9490 & 2.821 \\ 
HSDT~\cite{hsdt} & 0.522M & 855 & 28.508 & 0.9465 & 2.740 \\ 
VolFormer~\cite{yu2025volformer} & 29.195M & 41056 & 29.349 & 0.9345 & 2.508 \\ 
HIR-Diff~\cite{hir-diff}& 391.0M & 57567 & \textbf{30.123} & 0.9491 & 4.663 \\ 
\textbf{\ours{}} & 5.450M & 3720 & 30.099 & \textbf{0.9606} & \textbf{2.181} \\ 
\bottomrule
\end{tabular}}
\end{table}
\begin{figure*}[t]
\centering
\includegraphics[width=\textwidth]{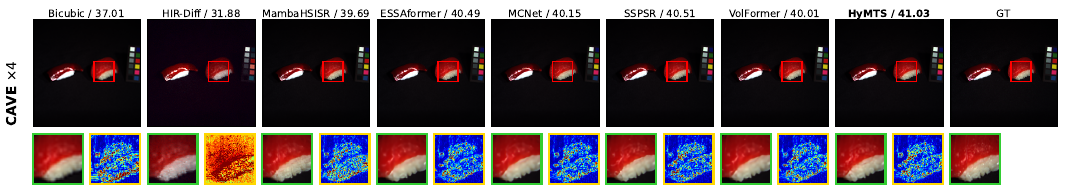}\\[-1pt]
{\small (a) CAVE, $\times4$ super-resolution}\\[4pt]
\includegraphics[width=\textwidth]{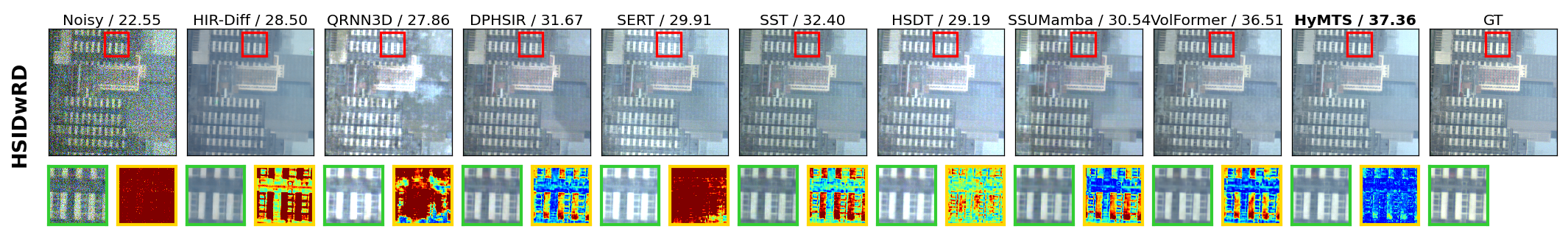}\\[-1pt]
{\small (b) HSIDwRD, real-noise denoising}
\caption{\textbf{Visual comparison.} (a) $\times4$ super-resolution on a CAVE test scene. (b) Real-noise denoising on an HSIDwRD test scene. Both are shown in bands 27, 15, and 5. In every image, the red box marks a region that is shown enlarged in the green box; the yellow box is the absolute error of the enlarged region, averaged over the bands (dark blue is zero, dark red is $0.03$ or more in (a) and $0.05$ or more in (b)). }
\label{fig:qual}
\end{figure*}
\subsection{Bidirectional Scan}
The bidirectional scan $\mathsf{S}$, Fig.~\ref{fig:schema}(c), aims to
link every pixel to its whole row and column with gated recurrent units
(GRUs)~\cite{rnngate}. With $\mathcal{Z}=\mathrm{LN}(\mathcal{F})$, row $i$
is read as the sequence $\mathbf{x}_t=\mathcal{Z}^{(:,i,t)}$,
$t=1,\dots,W$, by two GRUs with separate weights and states in
$\mathbb{R}^{C/2}$, one per direction, each starting from a zero state at
the image border,
\begin{equation}
\begin{aligned}
\overrightarrow{\mathbf{h}}_{t}&=\overrightarrow{\mathrm{GRU}}\big(\mathbf{x}_t,\overrightarrow{\mathbf{h}}_{t-1}\big),\\
\overleftarrow{\mathbf{h}}_{t}&=\overleftarrow{\mathrm{GRU}}\big(\mathbf{x}_t,\overleftarrow{\mathbf{h}}_{t+1}\big),\\
\mathcal{O}_{\mathrm{r}}^{(:,i,t)}&=[\overrightarrow{\mathbf{h}}_{t};\overleftarrow{\mathbf{h}}_{t}]\in\mathbb{R}^{C},
\end{aligned}
\label{eq:scan_row}
\end{equation}
so the output at a pixel depends on every pixel of its row. All rows
share this pair of GRUs; a second pair reads every column in the same way
and gives $\mathcal{O}_{\mathrm{c}}$. The two cubes are fused by a
$1\times1$ convolution with a residual,
\begin{equation}
\mathsf{S}(\mathcal{F})=\mathcal{F}+\mathbf{W}_\mathbf{mix}\,[\mathcal{O}_{\mathrm{r}};\mathcal{O}_{\mathrm{c}}],
\qquad \mathbf{W}_\mathbf{mix}\in\mathbb{R}^{C\times 2C},
\label{eq:scan}
\end{equation}
where $\mathbf{W}_\mathbf{mix}$ and its bias are zero-initialized. One scan links every pixel to its row and its column; the scan of the next block, acting on features that already carry these links, extends the dependency to the whole image (Sec.~\ref{sec:supp_scan}).
\subsection{Strip-window Coupling}
The strip-window layer $\mathsf{K}$ in Fig.~\ref{fig:schema}(d) considers each image row as a $C\times W$ plane. Hence, the plane $\mathcal{Z}^{(:,r,:)}\in\mathbb{R}^{C\times W}$ is the matrix whose $c$-th column is the feature vector $\mathcal{Z}^{(:,r,c)}$. $\Pi_W$ stacks the $H$ row planes into a batch of $H$ single-channel images of size $C\times W$, and $\Pi_W^{-1}$ restores the cube. A tensorial-summation map $\mathsf{M}_{C\times8}$ acts on every plane with windows of $C\times 8$ as shown in Fig.~\ref{fig:windows}(b), i.e.\ all $C$ channels of eight consecutive pixels. The plane $\mathcal{Z}^{(:,:,c)}\in\mathbb{R}^{C\times H}$ of $c$-th column works in the same way through $\Pi_H$ by the \emph{same} $\mathsf{M}_{C\times8}$ with the
same weights:
\begin{equation}
\begin{aligned}
\mathsf{K}(\mathcal{F})=\mathcal{F}
&+\Pi_W^{-1}\,\mathsf{M}_{C\times 8}\big(\Pi_W\,\mathrm{LN}(\mathcal{F})\big)\\
&+\Pi_H^{-1}\,\mathsf{M}_{C\times 8}\big(\Pi_H\,\mathrm{LN}(\mathcal{F})\big),
\end{aligned}
\label{eq:cs}
\end{equation}
where $\mathsf{M}_{C\times8}=\mathsf{T}'_{C\times8}\circ g\circ\mathsf{T}_{C\times8}$. In $\mathsf{T}_{C\times 8}$ the factors are $\mathbf{B}_t\in\mathbb{R}^{C\times C}$ and $\mathbf{D}_t\in\mathbb{R}^{8\times 8}$ with $T=3$. For each strip window that covers pixels $c_0+1,\dots,c_0+8$ of row $r$, the output is obtained by
\begin{equation}
\begin{aligned}
\mathcal{Z}'^{(:,\,r,\,c_0+j)}&=\sum_{i=1}^{8}\mathbf{M}_{ji}\,\mathcal{Z}^{(:,\,r,\,c_0+i)},\\
\boldsymbol{\Theta}_{ji}&=\sum_{t_0=1}^{T}\mathbf{D}_t^{(j,i)}\,\mathbf{B}_t\in\mathbb{R}^{C\times C},
\end{aligned}
\label{eq:cs_expand}
\end{equation}
where $j=1,\dots,8$ and $\boldsymbol{\Theta}_{ji}$ is channel-mixing matrix. The matrix
is from the $T$ shared channel factors $\mathbf{B}_t$, weighted by
the scalar $\mathbf{D}_t^{(j,i)}$ of the pixel pair, so the mixing of
channels depends on where the two pixels lie in the strip.
\subsection{Pooled Spectral Attention}
The pooled spectral attention $\mathsf{R}$, Fig.~\ref{fig:schema}(e), reweights the channels according to the content of the cube. Let $\mathcal{Z}=\mathrm{LN}(\mathcal{F})$ and let $\bar{\mathbf{z}}=\frac{1}{HW}\sum_{i,j}\mathcal{Z}^{(:,i,j)}\in\mathbb{R}^{C}$
be its spatial mean. $\mathsf{R}(\mathcal{F})$ is defined as:
\begin{equation}
\begin{aligned}
\relax[\mathbf{Q};\mathbf{K}]&=\mathbf{W}_{qk}\,\bar{\mathbf{z}},\qquad \mathcal{V}=\mathbf{W}_v\mathcal{Z},\\
\boldsymbol{\Phi}&=\operatorname{softmax}\!\Big(\frac{\mathbf{Q}\mathbf{K}^{\!\top}}{\sqrt{C}}\Big)\in\mathbb{R}^{C\times C},\\
\mathsf{R}(\mathcal{F})&=\mathcal{F}+\mathbf{W}_{p}\big(\boldsymbol{\Phi}\times_1\mathcal{V}\big),
\end{aligned}
\label{eq:sca}
\end{equation}
where $\mathbf{W}_{qk}\in\mathbb{R}^{2C\times C}$ and $\mathbf{W}_v,\mathbf{W}_p\in\mathbb{R}^{C\times C}$, and $\mathbf{Q},\mathbf{K}\in\mathbb{R}^{C}$ are the two halves of $\mathbf{W}_{qk}\bar{\mathbf{z}}$. The softmax is taken row-wise. $\boldsymbol{\Phi}$ is a $C\times C$ affinity and $\times_1$ is the mode-1 product of Eq.~\eqref{eq:ts}. Each row of $\boldsymbol{\Phi}$ is non-negative and sums to one, so every output channel is an adaptive average of the value channels. The affinity changes from cube to cube but not from pixel to pixel, which
distinguishes it from the SW-MSA (adaptive per pixel within its window) and
from the strip-window coupling (position-dependent but static). 
\subsection{Shifted-window Multi-head Self-attention}
The shifted-window multi-head self-attention (SW-MSA) is shown in Fig.~\ref{fig:windows}(c), We define $\mathcal{Z}=\mathrm{LN}(\mathcal{F})$ and $\mathsf{P}_a$ is the window partition of Eq.~\eqref{eq:part}. Each window of $\mathsf{P}_a\mathcal{Z}$ is flattened to a matrix $\mathbf{Z}\in\mathbb{R}^{M\times C}$ with $M=a^2$. Queries, keys, and values come from one bias-free $1\times1$ convolution shared by all windows of the layer, split among $h=4$ heads of dimension $d=C/h$,
\begin{equation}
\big[\mathbf{Q}_1\cdots\mathbf{Q}_h\,\big|\,\mathbf{K}_1\cdots\mathbf{K}_h\,\big|\,
\mathbf{V}_1\cdots\mathbf{V}_h\big]=\mathbf{Z}\,\mathbf{W}_{qkv},
\label{eq:qkv}
\end{equation}
with $\mathbf{W}_{qkv}\in\mathbb{R}^{C\times 3C}$ and
$\mathbf{Q}_i,\mathbf{K}_i,\mathbf{V}_i\in\mathbb{R}^{M\times d}$. Head $i$ mixes the $M$ pixels of the window with content-dependent weights,
\begin{equation}
\mathbf{O}_i=\operatorname{softmax}\!\Big(\frac{\mathbf{Q}_i\mathbf{K}_i^{\!\top}}{\sqrt{d}}
+\mathbf{B}^{\mathrm{rp}}_i+\mathbf{L}\Big)\mathbf{V}_i\in\mathbb{R}^{M\times d},
\label{eq:head}
\end{equation}
where $\mathbf{B}^{\mathrm{rp}}_i\in\mathbb{R}^{M\times M}$ is the relative-position bias of Eq.~\eqref{eq:rpb} and $\mathbf{L}\in\mathbb{R}^{M\times M}$ the shift mask of~\cite{swin2021}, zero in unshifted layers and in every interior window. The heads are concatenated,
$\mathrm{MSA}_a(\mathbf{Z})=[\mathbf{O}_1\cdots\mathbf{O}_h]\in\mathbb{R}^{M\times C}$. 

\begin{table*}[t]
\centering
\caption{Super-resolution on CAVE (24 training / 8 test scenes, whole $512\times512$ scenes), Chikusei (SSPSR split, four $512\times512$ test tiles) and Pavia Center (SSPSR split, four $223\times223$ test images). Best per dataset in bold (bicubic excluded).}
\label{tab:sr}\label{tab:cave}\label{tab:chikusei}\label{tab:pavia}
\footnotesize
\setlength{\tabcolsep}{2.5pt}
\resizebox{\textwidth}{!}{%
\begin{tabular}{@{}c l l l cccccc cccccc@{}}
\toprule
& & Params & GMACs & \multicolumn{6}{c}{$\times4$} & \multicolumn{6}{c}{$\times8$} \\
\cmidrule(lr){5-10}\cmidrule(lr){11-16}
Dataset & Method & ($\times4$\,/\,$\times8$) & ($\times4$\,/\,$\times8$) & MPSNR$\uparrow$ & MSSIM$\uparrow$ & SAM$\downarrow$ & CC$\uparrow$ & ERGAS$\downarrow$ & RMSE$\downarrow$ & MPSNR$\uparrow$ & MSSIM$\uparrow$ & SAM$\downarrow$ & CC$\uparrow$ & ERGAS$\downarrow$ & RMSE$\downarrow$ \\
\midrule
\multirow{9}{*}{\rotatebox[origin=c]{90}{CAVE}} & Bicubic & -- & -- & 35.549 & 0.9540 & 4.076 & 0.9880 & 6.195 & 0.0190 & 31.229 & 0.8964 & 5.274 & 0.9700 & 4.813 & 0.0307 \\ 
& MCNet~\cite{mcnet} & 2.174M / 2.961M & 1113 / 389 & 39.695 & 0.9791 & \textbf{3.303} & 0.9948 & 4.004 & 0.0119 & 34.896 & \textbf{0.9445} & \textbf{4.259} & \textbf{0.9845} & \textbf{3.370} & 0.0214 \\ 
& SSPSR~\cite{sspsr} & 26.077M / 28.438M & 2996 / 1803 & 39.726 & 0.9774 & 3.311 & 0.9946 & 4.049 & 0.0123 & 34.128 & 0.9335 & 4.312 & 0.9820 & 3.627 & 0.0233 \\ 
& ERCSR~\cite{ercsr} & 1.595M / 2.381M & 1090 / 598 & 39.633 & 0.9788 & 3.329 & 0.9946 & 4.070 & 0.0120 & 34.776 & 0.9420 & 4.326 & 0.9841 & 3.405 & 0.0216\\ 
& ESSAformer~\cite{essaformer} & 11.194M / 13.702M & 3273 / 3229 & 39.427 & 0.9776 & 3.452 & 0.9945 & 4.122 & 0.0124 & 34.494 & 0.9399 & 4.629 & 0.9835 & 3.462 & 0.0221 \\ 
& HIR-Diff~\cite{hir-diff} & 391.0M & 57567 & 31.514 & 0.8236 & 20.384 & 0.9716 & 10.011 & 0.0326 & 22.900 & 0.3654 & 25.592 & 0.8903 & 14.499 & 0.0951 \\ 
& VolFormer~\cite{yu2025volformer} & 29.470M / 29.617M & 3234 / 1677 & 39.791 & 0.9797 & 3.489 & 0.9950 & 3.910 & 0.0113 & 34.383 & 0.9402 & 5.293 & 0.9831 & 3.502 & 0.0219 \\ 
& MambaHSISR~\cite{mambahsisr} & 109.3M / 109.9M & 4346 / 1189 & 38.975 & 0.9742 & 3.558 & 0.9930 & 4.517 & 0.0138 & 34.290 & 0.9343 & 4.569 & 0.9816 & 3.631 & 0.0232 \\ 
& \textbf{\ours{}} & 11.121M / 11.465M & 316 / 200 & \textbf{40.427} & \textbf{0.9805} & 3.395 & \textbf{0.9954} & \textbf{3.702} & \textbf{0.0111} & \textbf{34.909} & 0.9427 & 4.347 & 0.9841 & 3.417 & \textbf{0.0212} \\ 
\midrule
\multirow{9}{*}{\rotatebox[origin=c]{90}{Chikusei}} & Bicubic & -- & -- & 37.638 & 0.8954 & 3.381 & 0.9212 & 6.756 & 0.0156 & 34.505 & 0.8069 & 5.037 & 0.8314 & 4.849 & 0.0224 \\ 
& MCNet~\cite{mcnet} & 2.174M / 2.961M & 4597 / 1605 & 38.426 & 0.9148 & 3.378 & 0.9310 & 6.309 & 0.0141 & 32.554 & 0.7326 & 7.071 & 0.7230 & 6.059 & 0.0283 \\ 
& SSPSR~\cite{sspsr} & 13.564M / 15.925M & 2783 / 1966 & 40.692 & 0.9480 & 2.250 & 0.9597 & 4.755 & 0.0110 & 36.029 & 0.8589 & 3.912 & 0.8825 & 4.062 & 0.0188 \\ 
& ERCSR~\cite{ercsr} & 1.595M / 2.381M & 4500 / 2470 & 39.323 & 0.9279 & 2.831 & 0.9454 & 5.547 & 0.0129 & 35.373 & 0.8335 & 4.538 & 0.8622& 4.369 & 0.0203 \\ 
& ESSAformer~\cite{essaformer} & 11.641M / 14.149M & 3336 / 3289 & 40.847 & 0.9500 & 2.217 & 0.9609 & 4.690 & 0.0108 & 35.908 & 0.8575 & 3.968 & 0.8784 & 4.145 & 0.0189 \\ 
& HIR-Diff~\cite{hir-diff} & 391.0M & 57567 & 33.030 & 0.7316 & 9.123 & 0.7939 & 12.329 & 0.0250 & 23.095 & 0.3199 & 21.880 & 0.3455 & 21.401 & 0.0945 \\ 
& VolFormer~\cite{yu2025volformer} & 30.140M / 30.288M & 10282 / 3740 & 39.751 & 0.9361 & 2.732 & 0.9501 & 5.352 & 0.0122 & 35.220 & 0.8339 & 4.811 & 0.8586 & 4.540 & 0.0204 \\ 
& MambaHSISR~\cite{mambahsisr} & 109.8M / 110.3M & 4423 / 1258 & 39.886 & 0.9390 & 2.423 & 0.9518 & 5.234 & 0.0120 & 35.202 & 0.8369 & 4.311 & 0.8568 & 4.491 & 0.0205 \\ 
& \textbf{\ours{}} & 8.672M / 8.707M & 266 / 158 & \textbf{40.957} & \textbf{0.9517} & \textbf{2.187} & \textbf{0.9618} & \textbf{4.638} & \textbf{0.0106} & \textbf{36.260} & \textbf{0.8656} & \textbf{3.766} & \textbf{0.8883} & \textbf{3.957} & \textbf{0.0182} \\ 
\midrule
\multirow{9}{*}{\rotatebox[origin=c]{90}{Pavia Center}} & Bicubic & -- & -- & 27.587 & 0.7218 & 6.135 & 0.8594 & 6.881 & 0.0437 & 24.597 & 0.4591 & 7.839 & 0.6970 & 4.828 & 0.0630 \\ 
& MCNet~\cite{mcnet} & 2.174M / 2.961M & 3663 / 1279 & 28.923 & 0.8043 & 5.894 & 0.8949 & 5.920 & 0.0373 & 25.031 & 0.5172 & 8.226 & 0.7244 & 4.592 & 0.0599 \\ 
& SSPSR~\cite{sspsr} & 13.385M / 15.745M & 2329 / 1668 & 28.444 & 0.7814 & 6.089 & 0.8821 & 6.259 & 0.0394 & 24.876 & 0.5029 & 8.417 & 0.7129 & 4.677 & 0.0606 \\ 
& ERCSR~\cite{ercsr} & 1.595M / 2.381M & 3586 / 1968 & 28.985 & 0.8079 & 5.927 & 0.8964 & 5.881 & 0.0370 & 24.968 & 0.5140 & 8.553 & 0.7203 & 4.626 & 0.0604 \\ 
& ESSAformer~\cite{essaformer} & 11.521M / 14.029M & 3319 / 3273 & 28.029 & 0.7601 & 7.058 & 0.8693 & 6.579 & 0.0412 & 23.943 & 0.4424 & 11.590 & 0.6396 & 5.229 & 0.0671 \\ 
& HIR-Diff~\cite{hir-diff} & 391.0M & 57567 & 27.179 & 0.7183 & 7.703 & 0.8366 & 7.424 & 0.0445 & 24.376 & 0.4295 & 9.203 & 0.6692 & 4.985 & 0.0634 \\ 
& VolFormer~\cite{yu2025volformer} & 29.960M / 30.108M & 8517 / 3221 & 28.847 & 0.8052 & 6.214 & 0.8927 & 5.984 & 0.0375 & 24.748 & 0.5024 & 8.990 & 0.7072 & 4.752 & 0.0615 \\ 
& MambaHSISR~\cite{mambahsisr} & 109.6M / 110.2M & 4402 / 1240 & 29.492 & 0.8328 & 5.221 & 0.9075 & 5.578 & 0.0347 & 24.824 & 0.5239 & 7.924 & 0.7140 & 4.711 & 0.0610 \\ 
& \textbf{\ours{}} & 6.505M / 6.606M & 215 / 135 & \textbf{29.632} & \textbf{0.8410} & \textbf{5.111} & \textbf{0.9105} & \textbf{5.491} & \textbf{0.0341} & \textbf{25.079} & \textbf{0.5409} & \textbf{7.250} & \textbf{0.7290} & \textbf{4.573} & \textbf{0.0593} \\ 
\bottomrule
\end{tabular}}
\end{table*}
\textbf{Relative-position bias and shifted windows.} We follow Swin~\cite{swin2021} to define $\mathbf{B}_i$ (Sec.~\ref{sec:supp_rpb}) and $\mathbf{L}$ (Sec.~\ref{sec:supp_shift}). Each head adds a learned bias $\mathbf{B}_i$ that depends only on the offset between the query and the key pixel. The SW-MSA alternates through the network depth between the window grid of Eq.~\eqref{eq:part} and the same grid displaced by half a window, $\delta=a/2$ pixels along both spatial axes. The displaced grid is obtained by rolling the features with $\mathrm{Roll}_{-\delta}$ and rolling the output back with $\mathrm{Roll}_{+\delta}$, so that consecutive attention layers exchange information across every window border.

The SW-MSA is then defined as:
\begin{equation}
\begin{aligned}
\mathsf{A}(\mathcal{F})&=\mathcal{F}+\mathbf{W}_{p}\,\mathrm{Roll}_{+\delta}
\mathsf{P}_a^{-1}\mathrm{MSA}_{a}\big(\mathsf{P}_a\mathrm{Roll}_{-\delta}\mathrm{LN}(\mathcal{F})\big),
\end{aligned}
\label{eq:wsa}
\end{equation}
where the output projection $\mathbf{W}_p$ is a $1\times1$ convolution with bias, both zero-initialized. $\delta$ = 0 in odd layers and $a / 2$ in even layers
\subsection{High-pass Gated Feed-forward Network}
A gated depth-wise feed-forward network (GDFN) ($\mathsf{G}$)~\cite{restormer2022} follows a SW-MSA ($\mathsf{A}$) as an attention unit to make it high-pass (HP-GDFN) by removing the spatial mean of its residual $\mathcal{Q}$. The GDFN residual is defined
$\mathcal{Q}$ is
\begin{equation}
\begin{aligned}
\mathcal{Q}=\mathbf{W}_o\Big(&\phi\big(\mathrm{DW}_1 \mathbf{W}_1\,\mathrm{LN}(\mathcal{F})\big)\\
&\odot\mathrm{DW}_2 \mathbf{W}_2\,\mathrm{LN}(\mathcal{F})\Big),\qquad \mathbf{W}_o\leftarrow 0,
\end{aligned}
\label{eq:gdfn_res}
\end{equation}
where $\mathbf{W}_1,\mathbf{W}_2$ map $C$ to $C$ channels (expansion ratio one) and $\mathbf{W}_o$ has no bias. $\phi$ is GELU, and $\mathrm{DW}_1$, $\mathrm{DW}_2$ are $3 \times 3$ depth-wise convolution.We add this residual after removing its spatial mean in every channel as:
\begin{equation}
\mathsf{G}(\mathcal{F})=\mathcal{F}+(\mathrm{I}-\mu)\,\mathcal{Q},
\label{eq:gdfn}
\end{equation}
where $\mu$ replaces every entry of channel $c$ by the channel mean
$\frac{1}{HW}\sum_{i,j}\mathcal{Q}^{(c,i,j)}$ over all $H\times W$
positions of the input,
the training patch during training and the test frame or tile at
inference; $\mu$ has no parameters and no window size. 

\subsection{Training objectives}
\label{sec:train}
\textbf{Denoising.} With $\tilde{\mathcal{Y}}$ the output and
$\mathcal{Y}$ the ground truth, the denoiser is trained on Gaussian noise
with
\begin{equation}
\mathcal{L}_{\mathrm{dn}}=\|\tilde{\mathcal{Y}}-\mathcal{Y}\|_2^2+\lambda_{\mathrm{ssim}}\big(1-\mathrm{SSIM}(\tilde{\mathcal{Y}},\mathcal{Y})\big),
\label{eq:loss_dn}
\end{equation}
and on real noise with the $\ell_1$ distance
$\|\tilde{\mathcal{Y}}-\mathcal{Y}\|_1$.

\textbf{Super-resolution.} The loss is a band-weighted $\ell_1$ distance
plus a spectral total-variation term on the output,
\begin{equation}
\begin{aligned}
\mathcal{L}_{\mathrm{sr}}=\sum_{b=1}^{C_b} w_b\,\|\tilde{\mathbf{Y}}_b-\mathbf{Y}_b\|_1
&+\lambda_{\mathrm{tv}}\sum_{b=1}^{C_b-1}\|\tilde{\mathbf{Y}}_{b+1}-\tilde{\mathbf{Y}}_b\|_1 ,\\
w_b&\propto\frac{1}{\bar e_b +\epsilon},
\end{aligned}
\label{eq:loss_sr}
\end{equation}
where $\bar e_b$ is the running root-mean-square error of band $b$. Low-energy bands have small absolute errors, so the inverse-error weights keep them from being ignored. The weights $\lambda_{\mathrm{s}}$ and $\lambda_{\mathrm{tv}}$ and the settings of each dataset are given in Sec.~\ref{sec:supp_train}.

\section{Experiments}
\label{sec:experiments}
\subsection{Datasets and evaluation metrics}
\textbf{Denoising.} We implement HSI denoising on ICVL~\cite{icvl_2016} database with Gaussian noise and the real noise database HSIDwRD~\cite{hsidwrd2021}. The ICVL contains 201 indoor and outdoor scenes with $1392 \times 1300$ pixels and 31 bands. We use 100 scenes for training and 50 for testing, as in the QRNN3D/SERT split~\cite{3dqrnn, sert}. We normalize each scene by its maximum value. We use the same cropping methods as in QRNN3D/SERT~\cite{3dqrnn, sert} to crop training and testing scenes. The noise added for all models (HyMTS and baselines) during training is i.i.d.\ Gaussian with $\sigma\in\{10,30,50,70\}$ on the 0-255 scale. The blind setting during testing is $\sigma\sim\mathcal{U}[30,70]$ per image. HSIDwRD~\cite{hsidwrd2021} is a database with real noise, which contains 59 scenes of $696\times520$ pixels and 34 bands. We follow SERT's partition of 44 training and 15 test scenes and its input pipeline, which divides both cubes by 4096 and multiplies the noisy one by the exposure ratio of 50. Test inputs are the center $512\times512$ crop of each test scene, matching SERT's noisy input scores of 23.26\,dB.

\noindent\textbf{Super-resolution.} We implement super-resolution on CAVE~\cite{cave}, Chikusei~\cite{chikukei} and Pavia~\cite{paiva}. CAVE contains 32 scenes with $512 \times 512 \times 31$. We train all models with the same 24 scenes and use the remaining 8 scenes for testing. The low-resolution inputs are produced by bicubic downsampling. Chikusei~\cite{chikukei} is captured by the Hyperspec-VNIR-CIRIS spectrometer with $ 2517 \times 2335$ pixels and 128 bands. In this work, we follow SSPSR~\cite{sspsr} to crop the raw scene to $2304 \times 2048 \times 128$ around the center region. From the cropped scene, we cut off the top 512 rows to create four $512 \times 512$ test tiles, and use the remaining $1792 \times 2048$ region for training. Pavia Centre~\cite{paiva} contains 115 bands and $1096 \times 1096$ pixels. We keep the effective region and bands as $1096 \times 715 \times 102$. As in SSPSR~\cite{sspsr}, we crop four non-overlapping $223 \times 223$ test images from the left part, and training patches from the rest. 

\subsection{Training Settings.}
All models are trained from scratch on a single AMD MI250X graphics with AdamW ($\beta=(0.9,0.9)$, weight decay $10^{-4}$). The initial learning rate is $2\times10^{-4}$, annealed by a cosine schedule to $10^{-6}$ over the full run, with gradient-norm clipping. Training patches are augmented with random horizontal and vertical flips and $90^\circ$ rotations. We keep an exponential moving average of the weights with decay 0.999 on ICVL, HSIDwRD, Chikusei, and Pavia. We evaluate the final iterate of each run: the averaged weights on ICVL, Chikusei and Pavia, and the raw weights on HSIDwRD and CAVE. Other details are described in the supplementary in Sec.~\ref {sec:supp_train}.
\subsection{Experimental Results Analysis}
\textbf{Gaussian denoising.} On ICVL (Tab.~\ref{tab:icvl}), HyMTS has the highest MPSNR at every noise level: $0.06$ to $0.15$\,dB above SSUMamba, the next best, and at $\sigma{=}50$ $0.20$\,dB above SERT and $0.94$\,dB above VolFormer. MSSIM is within $10^{-4}$ of SST, SERT, SSUMamba, and HyMTS. \ours{} has achieved the best SAM except at $\sigma{=}10$, where SSUMamba does. It uses $3720$ GMACs, $2.6$ times fewer than SSUMamba and $9.2$ times fewer than VolFormer, but $3.1$ and $6.9$ times more than SST and SERT. The training-free HIR-Diff is $14$\,dB below the trained networks at $\sigma{=}50$.

\noindent\textbf{Super-resolution.} HyMTS has achieved $0.64$\,dB above VolFormer and $0.70$\,dB above SSPSR at $\times4$ without RGB co-training On CAVE (Tab.~\ref{tab:cave}). It is $0.01$\,dB above MCNet and $0.41$\,dB above VolFormer at $\times8$. On Chikusei (Tab.~\ref{tab:chikusei}), it is $0.11$\,dB above ESSAformer at $\times4$ and $0.23$\,dB above SSPSR at $\times8$. On Pavia Centre (Tab.~\ref{tab:pavia}) \ours{} is $0.14$\,dB above MambaHSISR and $0.65$\,dB above ERCSR at $\times4$, and $0.05$\,dB above MCNet and $0.11$\,dB above ERCSR at $\times8$, where no network gains more than $0.48$\,dB over bicubic interpolation. HyMTS achieves the best value on all six metrics on Chikusei and Pavia Center at both scales. \ours{} needs $135$ to $316$ GMACs, $5.9$ to $39.6$ times fewer than ESSAformer, VolFormer and MambaHSISR and $1.9$ to $17.3$ times fewer than SSPSR, MCNet and ERCSR.

\noindent\textbf{Visual results.} 
Fig.~\ref{fig:qual} shows a CAVE scene at $\times4$ and an HSIDwRD scene with real noise. In (a), all trained networks sharpen the material boundary that bicubic interpolation blurs, and \ours{} has the lowest error. Its MPSNR on this scene is the highest, $0.52$\,dB above SSPSR, the next best. In (b), the other methods leave errors on the bright window panels or over the whole region, while \ours{} recovers both. Its MPSNR on this scene is $37.36$\,dB, $0.85$\,dB above VolFormer and $4.96$\,dB above SST, the next two methods. Other visual results are shown in Sec.~\ref{sec:supp_visual}.

\noindent\textbf{Ablation Studies}
We conduct ablations to assess the importance of components and
exposure calibration on HSIDwRD, as reported in Sec.~\ref{sec:op_result}. The component ablation removes or replaces one component at a time, and the exposure calibration retrains \ours{} and four baselines with the exposure ratio fitted per training scene.

\section{Conclusion}
\label{sec:conclusion}
We introduce HyMTS, a novel network for hyperspectral image restoration. Rather than relying mainly on attention mechanisms, HyMTS models local spectral--spatial correlations with factorized multi-scale projections. These projections cover multi-scale
spectral-spatial windows with few parameters per receptive field, which keeps the model compact. The input-adaptive layers are reserved for longer-range context, then local and global dependencies are captured within one block. The proposed HyMTS backbone can be used for different degradations as both a denoiser and a super-resolution network. Extensive experiments on five HSI benchmarks, covering synthetic noise, real noise, and two super-resolution scales, demonstrate HyMTS's superior robustness and effectiveness under different degradation conditions.

{
    \small
    \bibliographystyle{ieeenat_fullname}
    \bibliography{main}
}

\section{Components of one \ours{} block}
\label{sec:supp_components}
Table~\ref{tab:components} lists the components of one \ours{} block with
their class, extent, and parameter share.
\begin{table}[h]
\centering
\footnotesize
\setlength{\tabcolsep}{3pt}
\caption{\textbf{Components of one \ours{} block}, Shares are the fraction of the parameters held
by all instances of a component in the ICVL denoiser ($5.450$\,M) and the
Chikusei $\times 4$ network ($8.672$\,M).}
\label{tab:components}
\begin{tabular}{@{}llllrr@{}}
\toprule
& Component & Class & Extent & \multicolumn{2}{c}{Share (\%)}\\
\cmidrule(lr){5-6}
& & & & DN & SR\\
\midrule
$\mathsf{B}$ & Tensorial summation & static & $w\times w$, $w\in\Omega$ & 20.0 & 44.8\\
$\mathsf{K}$ & Strip-window coupling & static & $8$-px strip, all bands & 2.1 & 7.0\\
$\mathsf{S}$ & Bidirectional scan & gated & whole row, column & 3.8 & 12.9\\
$\mathsf{G}$ & HP-GDFN & gated & $3\times 3$ & 26.0 & 11.0\\
$\mathsf{R}$ & Pooled spectral attention & adaptive & all bands, pooled & 1.4 & 4.6\\
$\mathsf{A}$ & SW-MSA & adaptive & $a\times a$, shifted & 46.7 & 15.0\\
\bottomrule
\end{tabular}
\end{table}

\section{Relative-position bias}
\label{sec:supp_rpb}

The scores
$\mathbf{Q}_i\mathbf{K}_i^{\!\top}$ in Eq.~\eqref{eq:head} depend only on
the feature vectors of two pixels, not on where the pixels lie in the
window. We therefore add the relative-position bias of
Swin~\cite{swin2021}: a learned scalar for each offset between a query
pixel $m=(p,q)$ and a key pixel $m'=(p',q')$ of the window. Both offset
components $\Delta p=p-p'$ and $\Delta q=q-q'$ lie in
$\{-(a-1),\dots,a-1\}$, so head $i$ needs only a vector
$\mathbf{b}_i\in\mathbb{R}^{(2a-1)^2}$, from which
\begin{equation}
\begin{aligned}
\mathbf{B}_i^{(m,m')}&=\mathbf{b}_i^{(\iota(m,m'))},\\
\iota(m,m')&=(\Delta p+a-1)(2a-1)+(\Delta q+a-1).
\end{aligned}
\label{eq:rpb}
\end{equation}
Each SW-MSA has its own $h=4$ vectors and uses them in every window.
They are initialized from a truncated normal distribution with standard deviation $0.02$. At $a=32$ they hold $4\cdot63^2=15.9$\,k parameters, $30\%$ of the SW-MSA.
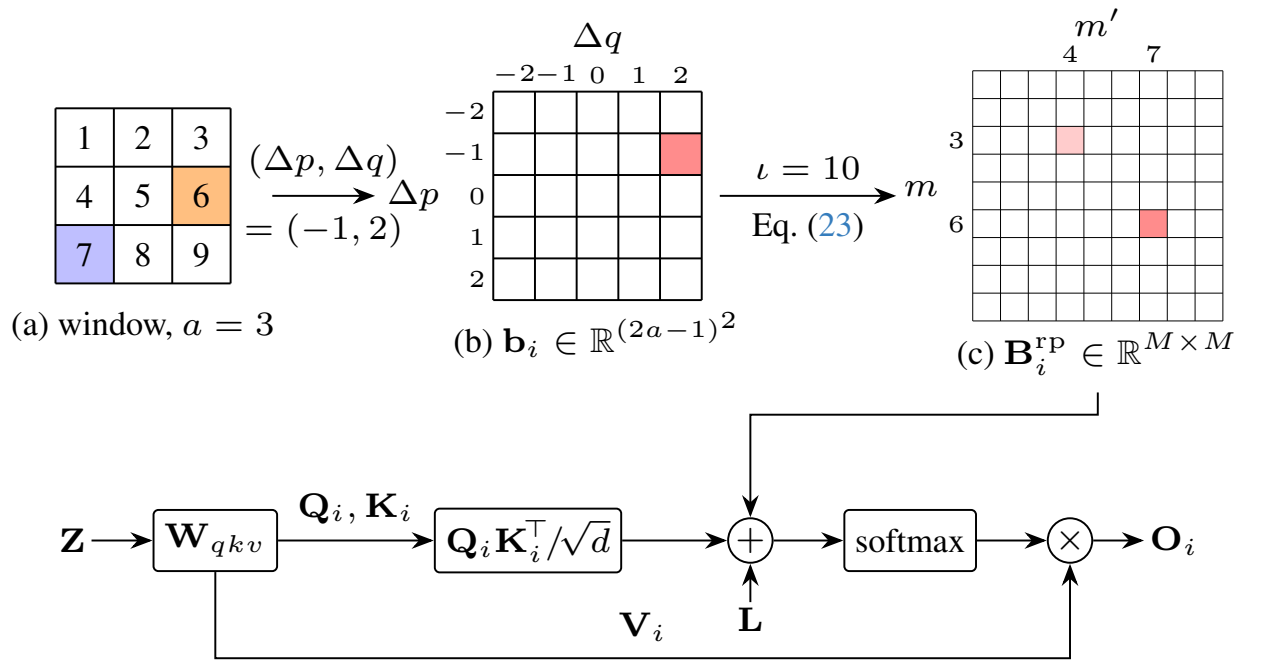
\begin{figure}[t]
\centering
\resizebox{\columnwidth}{!}{%
\begin{tikzpicture}[font=\scriptsize,>={Stealth[length=1.6mm]},
  op/.style={draw,rounded corners=1pt,inner sep=2.2pt,minimum height=4.4mm},
  jn/.style={draw,circle,inner sep=0pt,minimum size=3.2mm}]
\begin{scope}
\fill[orange!50] (0.84,0.42) rectangle (1.26,0.84);
\fill[blue!25] (0,0) rectangle (0.42,0.42);
\foreach \i in {0,...,3}{
  \draw (0.42*\i,0)--(0.42*\i,1.26);
  \draw (0,0.42*\i)--(1.26,0.42*\i);}
\foreach \p in {1,2,3}{\foreach \q in {1,2,3}{
  \pgfmathtruncatemacro{\n}{(\p-1)*3+\q}
  \node[inner sep=0pt] at (0.42*\q-0.21,1.47-0.42*\p) {\n};}}
\node[anchor=north] at (0.63,-0.08) {(a) window, $a=3$};
\end{scope}
\begin{scope}[shift={(3.15,-0.12)}]
\fill[red!45] (1.2,0.9) rectangle (1.5,1.2);
\foreach \i in {0,...,5}{
  \draw (0.3*\i,0)--(0.3*\i,1.5);
  \draw (0,0.3*\i)--(1.5,0.3*\i);}
\foreach \v [count=\j] in {-2,-1,0,1,2}{
  \node[font=\tiny,inner sep=0pt] at (0.3*\j-0.15,1.62) {$\v$};
  \node[font=\tiny,inner sep=0pt,anchor=east] at (-0.05,1.65-0.3*\j) {$\v$};}
\node[inner sep=0pt] at (0.75,1.88) {$\Delta q$};
\node[inner sep=0pt,anchor=east] at (-0.38,0.75) {$\Delta p$};
\node[anchor=north] at (0.75,0.04) {(b) $\mathbf{b}_i\in\mathbb{R}^{(2a-1)^2}$};
\end{scope}
\begin{scope}[shift={(6.6,-0.27)}]
\fill[red!45] (1.2,0.6) rectangle (1.4,0.8);
\fill[red!20] (0.6,1.2) rectangle (0.8,1.4);
\foreach \i in {0,...,9}{
  \draw[very thin] (0.2*\i,0)--(0.2*\i,1.8);
  \draw[very thin] (0,0.2*\i)--(1.8,0.2*\i);}
\node[font=\tiny,inner sep=0pt] at (1.3,1.92) {$7$};
\node[font=\tiny,inner sep=0pt] at (0.7,1.92) {$4$};
\node[font=\tiny,inner sep=0pt,anchor=east] at (-0.05,0.7) {$6$};
\node[font=\tiny,inner sep=0pt,anchor=east] at (-0.05,1.3) {$3$};
\node[inner sep=0pt] at (0.9,2.15) {$m'$};
\node[inner sep=0pt,anchor=east] at (-0.24,0.95) {$m$};
\node[anchor=north] (Bt) at (0.9,0.04) {(c) $\mathbf{B}^{\mathrm{rp}}_i\in\mathbb{R}^{M\times M}$};
\end{scope}
\draw[->] (1.56,0.63)--node[above]{$(\Delta p,\Delta q)$}node[below]{$=(-1,2)$}(2.30,0.63);
\draw[->] (4.78,0.63)--node[above]{$\iota=10$}node[below]{Eq.~\eqref{eq:rpb}}(6.05,0.63);
\node[inner sep=1pt] (Z) at (0.12,-1.85) {$\mathbf{Z}$};
\node[op] (W) at (1.15,-1.85) {$\mathbf{W}_{qkv}$};
\node[op] (S) at (3.4,-1.85) {$\mathbf{Q}_i\mathbf{K}_i^{\!\top}\!/\!\sqrt{d}$};
\node[jn] (P) at (5.0,-1.85) {$+$};
\node[op] (SM) at (6.15,-1.85) {softmax};
\node[jn] (X) at (7.3,-1.85) {$\times$};
\node[inner sep=1pt] (O) at (8.05,-1.85) {$\mathbf{O}_i$};
\node[inner sep=1pt] (L) at (5.0,-2.42) {$\mathbf{L}$};
\draw[->] (Z)--(W);
\draw[->] (W)--node[above]{$\mathbf{Q}_i,\mathbf{K}_i$}(S);
\draw[->] (S)--(P);
\draw[->] (P)--(SM);
\draw[->] (SM)--(X);
\draw[->] (X)--(O);
\draw[->] (L)--(P);
\draw[->] (W.south)--++(0,-0.62)-|node[pos=0.25,above]{$\mathbf{V}_i$}(X.south);
\draw[->] (Bt.south)--++(0,-0.16)-|(P.north);
\end{tikzpicture}}
\caption{\textbf{Data flow of the relative-position bias}, $a=3$, $M=9$.
(a) The query $m=(2,3)$, pixel $6$, and the key $m'=(3,1)$, pixel $7$,
have the offset $(\Delta p,\Delta q)=(-1,2)$. (b) The offset addresses
entry $\iota=10$ of $\mathbf{b}_i$, drawn as a $(2a-1)\times(2a-1)$ table. (c) The entry is copied to position $(6,7)$ of $\mathbf{B}^{\mathrm{rp}}_i$ and to every other pair with the same offset, here $(3,4)$. The matrix is added to the scores of Eq.~\eqref{eq:head} before the softmax.}
\label{fig:rpb}
\end{figure}

\section{Shifted windows}
\label{sec:supp_shift}
As in Swin~\cite{swin2021}, the SW-MSA are numbered $\ell=1,\dots,Nk$ through the network ($N$ blocks
of $k$ attention units), and the $\ell$-th is within the
windows of $\mathsf{P}_a\mathrm{Roll}_{-\delta}\mathcal{Z}$,
\begin{equation}
\begin{aligned}
\mathcal{W}^{(c,p,q)}&=\mathcal{Z}^{(c,\;\langle ua+p+\delta\rangle_H,\;\langle va+q+\delta\rangle_W)},\\
\delta&=\begin{cases}0, & \ell\ \text{odd},\\ a/2, & \ell\ \text{even},\end{cases}
\end{aligned}
\label{eq:roll}
\end{equation}
where $\langle i\rangle_H=((i-1)\bmod H)+1$: the windows of
Eq.~\eqref{eq:part} displaced by $\delta$ on both axes and wrapped at the
image border. A wrapped window joins pixels from opposite edges of the
image, which the mask of Eq.~\eqref{eq:head} separates,
\begin{equation}
\begin{aligned}
\mathbf{L}^{(m,m')}&=\begin{cases}
0, & r(m)=r(m'),\\
-100, & r(m)\neq r(m'),
\end{cases}\\
r(m)&=\big(\mathbf{1}[ua+p+\delta>H],\;\mathbf{1}[va+q+\delta>W]\big),
\end{aligned}
\label{eq:mask}
\end{equation}
with $m=(p,q)$ and $\mathbf{1}[\cdot]$ the indicator; $\mathbf{L}=\mathbf{0}$ for $\delta=0$ and in every
interior window. $\mathrm{Roll}_{+\delta}$ returns the output of each
window to its pixels Eq.~\eqref{eq:wsa}. Every border of the unshifted
grid then lies at the center of a shifted window, and the receptive field
of a stack of attention units grows by $a/2$ on every side per layer. It
covers the whole frame after $\max(H,W)/a$ layers, $16$ for a
$512\times512$ ICVL crop at $a=32$. The shift is applied whenever the padded extent
exceeds $a$ in both axes. Only the attention windows are shifted; the tensorial-summation
grid stays anchored at the origin in every reported network.

\section{Bidirectional scan}
\label{sec:supp_scan}
\textbf{Recurrence.} Along a row, the gated recurrent unit~\cite{rnngate} of Eq.~\eqref{eq:scan_row} reads at pixel $t$ the feature vector $\mathbf{x}_t$ and the state
$\mathbf{h}_{t-1}\in\mathbb{R}^{C/2}$ carried from the previous pixel; the
state is a running summary of the pixels already visited. The update gate
$\mathbf{z}_t\in(0,1)^{C/2}$, a sigmoid of $\mathbf{x}_t$ and
$\mathbf{h}_{t-1}$, decides, channel by channel, how much of the summary
to keep and how much to replace with a candidate $\tilde{\mathbf{h}}_t$
formed from the current feature vector,
\begin{equation}
\mathbf{h}_t=(\mathbf{1}-\mathbf{z}_t)\odot\mathbf{h}_{t-1}+\mathbf{z}_t\odot\tilde{\mathbf{h}}_t,
\label{eq:gru_update}
\end{equation}

\section{Network configures}
\label{sec:trunk}
Equation~\eqref{eq:cycle} has two kinds of settings. The number of tensorial-summation layers $n_b=2N$ and the number of attention units $k$ per HyMTS block. Tab.~\ref{tab:icvl_ablate} uses different settings, moving from $n_b=12$, $k=1$ to $n_b=4$, $k=24$.

\noindent\textbf{Denoising.} We set $C=96$, $n_b=4$ ($N=2$), $T=3$, $\Omega=\{4,8,16\}$,
attention window $a=32$ and $k=24$: $5.450$\,M parameters for the $31$
bands of ICVL and the $34$ bands of HSIDwRD. The SW-MSA occupies $46.7\%$ of the
parameters (of which the relative-position biases are $14.0\%$), the
HP-GDFNs take $26.0\%$, the tensorial-summation layers use $20.0\%$, the
scans use $3.8\%$, the strip-window layers use $2.1\%$, and the pooled spectral
attention uses $1.4\%$. The same network serves every noise level.

\noindent\textbf{Super-resolution.} We use $n_b=16$ ($N=8$), $k=3$, $T=1$ in the backbone and $T=3$ in the strip-window and upsampling maps, the shrinkage of Eq.~\eqref{eq:block}, and $\Omega=\{4,8,16,32\}$ with $a=16$ at $\times 4$ and $\Omega=\{4,8,16\}$
with $a=8$ at $\times 8$, so that the attention window covers the
same high-resolution extent. The width is $C=96$ on Pavia, $C=112$ on Chikusei, and $C=128$ on CAVE. In the Chikusei $\times 4$ network, the
tensorial-summation layers hold $44.8\%$ of the parameters, the SW-MSA
$15.0\%$, the scans $12.9\%$, the HP-GDFNs $11.0\%$, the
strip-window layers $7.0\%$, the pooled spectral attention $4.6\%$ and the
whole super-resolution head $4.4\%$. The $\times 8$ and $\times 4$ networks are trained from scratch. Pavia $\times 8$ is trained for $10$k iterations because its training region is small, with the same backbone without the PixelShuffle path.

\section{Training details and Metrics per dataset}
\label{sec:supp_train}
We train all networks from scratch with AdamW, a cosine schedule, an exponential moving average of the weights~\cite{polyak1992} with decay $0.999$, and the per-dataset settings in Tab.~\ref{tab:train_settings}. Tab.~\ref{tab:split_indices} shows the training and test splits. We report the final iterate of every run. No auxiliary RGB data is used in any experiments. We evaluate denoising experiments using MPSNR, MSSIM, and SAM~\cite{yu2025volformer,sspsr}. We evaluate super-resolution using six metrics from~\cite {yu2025volformer,sspsr}: MPSNR, MSSIM, SAM, CC, ERGAS, and RMSE, with ERGAS at the true scale factor. GMACs are multiply-accumulate operations for one $512\times512$ cube. 

\noindent\textbf{Denoising.} One blind model is used all noise levels. On ICVL, it is
trained on $64\times 64$ patches with Gaussian noise $\sigma\in[10,70]$ and the loss $\mathcal{L}_{\mathrm{dn}}$ of Eq.~\eqref{eq:loss_dn} with $\lambda_{\mathrm{s}}=5\times 10^{-3}$. On HSIDwRD, the model is trained on $128\times 128$ patches of the real short-exposure/long-exposure pairs with the $\ell_1$ loss.
\begin{table*}[t]
\centering
\caption{Image size, bands, training and test splits and training settings of HyMTS per dataset. Width is the number of feature channels $C$ in the backbone. Patch sizes are high-resolution pixels.}
\label{tab:train_settings}
\small
\setlength{\tabcolsep}{4pt}
\begin{tabular}{@{}lccccc@{}}
\toprule
 & ICVL & HSIDwRD & Chikusei & CAVE & Pavia \\
\midrule
Image size & $1392\times1300$ & $696\times520$ & $2304\times2048$ & $512\times512$ & $1096\times715$ \\
Bands & 31 & 34 & 128 & 31 & 102 \\
Training set & 100 scenes & 44 scenes & bottom $1792\times2048$ & 24 scenes & right $1096\times492$ \\ 
Test set & 50 scenes & 15 scenes & 4 tiles, top 512 rows & 8 scenes & 4 images, left 223 columns \\
Training patch & $64^2$ & $128^2$ & $128^2$ & $128^2$ & $128^2$ \\ 
Test input & centre $512^2$ crop & centre $512^2$ crop & whole $512^2$ tile & whole $512^2$ scene & whole $223^2$ image \\
\midrule
Width $C$ & 96 & 96 & 112 & 128 & 96 \\
Batch & 4 & 2 & 8 & 8 & 8 \\
Iterations ($\times4$/$\times8$) & 150k & 36k & 50k/50k & 75k/150k & 10k/10k \\
FFN dropout & -- & -- & 0.2 & 0.2 & 0.2 \\
Weight EMA & 0.999 & 0.999 & 0.999 & 0.999 & 0.999 \\
Grad-clip norm & 0.01 & 0.01 & 1.0 & 1.0 & 1.0 \\
\bottomrule
\end{tabular}
\end{table*}

\noindent\textbf{Super-resolution.} In $\mathcal{L}_{\mathrm{sr}}$ of
Eq.~\eqref{eq:loss_sr}, $\bar\varepsilon_b$ is an exponential moving
average (decay $0.9$) of
the root-mean-square error of band $b$ over the training batches, detached
from the graph, $\epsilon=10^{-3}$, and the weights are normalised to mean
one over bands. The band weighting is used on Chikusei, CAVE and Pavia $\times 8$ and
$\lambda_{\mathrm{tv}}=10^{-2}$ on Chikusei $\times 4$ and Pavia $\times 4$; CAVE
$\times 4$ adds a $(1-\mathrm{SSIM})$ term with weight $0.2$. Code,
configurations and checkpoints will be released.
\begin{table*}[t]
\centering
\caption{Training and test split of each dataset. Pixel indices are
1-based and inclusive. The Chikusei image is rows 107--2410 and columns
144--2191 of the released scene. The ICVL and HSIDwRD splits are those of~\cite{sert}; the CAVE test scenes are indices 2, 10, 11, 14, 16, 22, 26 and 28 of the alphabetical list, drawn
with the released seed of ESSAformer.}
\label{tab:split_indices}
\footnotesize
\setlength{\tabcolsep}{4pt}
\begin{tabular}{@{}l p{0.27\textwidth} p{0.58\textwidth}@{}}
\toprule
Dataset & Training & Test \\
\midrule
ICVL & the 100 scenes of the training list of~\cite{sert} & \raggedright CC\_40D\_2\_1103-0917, IDS\_COLORCHECK\_1020-1215-1, Labtest\_0910-1504, Labtest\_0910-1509, Labtest\_0910-1513, Lehavim\_0910-1636, Lehavim\_0910-1716, Lehavim\_0910-1717, Master5000K\_2900K, bulb\_0822-0909, gavyam\_0823-0930, gavyam\_0823-0933, gavyam\_0823-0944, lehavim\_0910-1600, lehavim\_0910-1602, lehavim\_0910-1610, nachal\_0823-1047, nachal\_0823-1118, nachal\_0823-1121, nachal\_0823-1127, nachal\_0823-1144, nachal\_0823-1149, nachal\_0823-1214, nachal\_0823-1217, nachal\_0823-1222, nachal\_0823-1223, negev\_0823-1003, objects\_0924-1650, rmt\_0328-1241-1, rmt\_0328-1249-1, selfie\_0822-0906, ulm\_0328-1118, Lehavim\_0910-1626, Lehavim\_0910-1635, Master2900k, Master5000K, objects\_0924-1550, objects\_0924-1623, objects\_0924-1557, objects\_0924-1602, objects\_0924-1607, objects\_0924-1611, objects\_0924-1614, objects\_0924-1617, objects\_0924-1625, objects\_0924-1628, objects\_0924-1632, objects\_0924-1637, objects\_0924-1641, objects\_0924-1648 \tabularnewline
\addlinespace
HSIDwRD & scenes 11--50 and 56--59 & scenes 1--10 and 51--55 \\
\addlinespace
CAVE & the other 24 scenes & \raggedright cd, fake and real lemon slices, fake and real lemons, fake and real sushi, feathers, paints, real and fake peppers, stuffed toys \tabularnewline
\addlinespace
Chikusei & rows 513--2304, all columns & rows 1--512; four tiles at columns 1--512, 513--1024, 1025--1536 and 1537--2048 \\
\addlinespace
Pavia Centre & columns 224--715, all rows & columns 1--223; four images at rows 1--223, 224--446, 447--669 and 670--892 (rows 893--1096 unused) \\
\bottomrule
\end{tabular}
\end{table*}

\section{Ablation studies}
\label{sec:op_result}
\textbf{Components.} We conduct experiments to investigate the importance of one component on ICVL and HSIDwRD in Tab.~\ref{tab:icvl_ablate} and Tab.~\ref{tab:hsidwrd_ablate}. Without the window shift of the SW-MSA, MPSNR at $\sigma{=}50$ drops by $0.30$\,dB. Moving the budget from attention to multilinear depth ($N{=}6$, $k{=}1$, six attention units instead of 48) costs $0.68$\,dB. Without the strip-window coupling, training peaks at 18k iterations and degrades afterward, to $7.55$\,dB below the full model at the last iterate. Replacing the tensorial-summation maps with an iso-parameter dense MLP changes MPSNR by $0.08$\,dB.

Tab.~\ref{tab:hsidwrd_ablate} presents the component ablation on HSIDwRD. Without the strip-window coupling, the MPSNR decreases by $0.47$\,dB. Replacing the dense MLP reduces the MPSNR by $1.96$\,dB. The scan $\mathsf{S}$ and pooled spectral attention $\mathsf{R}$ can improve the MPSNR by $1.72$ and $1.22$\,dB, respectively. On Pavia Center $\times4$ (Tab.~\ref{tab:pavia_ablate}), removing the window shift and
reducing the attention to $k{=}1$ change MPSNR by $-0.09$ and
$+0.09$\,dB.

\noindent\textbf{Classes.} Tab.~\ref{tab:class_ablate} removes all components of one class in Tab.~\ref{tab:components} from the backbone: the static ($\mathsf{B}$, $\mathsf{K}$), the gated ($\mathsf{S}$, $\mathsf{G}$) or the adaptive ($\mathsf{R}$, $\mathsf{A}$) components. On ICVL, both the static and adaptive components contribute heavily: without either, performance degrades to $13.52$ and $8.64$\,dB below the full model at the last iterate. On HSIDwRD every class contributes, from $0.60$\,dB (static) to $2.04$\,dB (adaptive). On Pavia Center $\times4$, the gated class is the most important, as without it MPSNR falls by $0.67$\,dB.

\noindent\textbf{Exposure calibration.} HSI pairs in HSIDwRD contain a short exposure with a long one, and the short exposure is amplified by the nominal ratio of $50$. The true ratio differs between scenes. We measure it by the least-squares gain $g$ of the long exposure $\mathcal{Y}$ on the amplified short exposure $50\,\hat{\mathcal{Y}}$. It is computed over the unsaturated pixels as follows: 
\begin{equation}
g=\frac{\langle\mathcal{Y},\,50\,\hat{\mathcal{Y}}\rangle}{\langle 50\,\hat{\mathcal{Y}},\,50\,\hat{\mathcal{Y}}\rangle},
\label{eq:calib}
\end{equation}
Over the $44$ training scenes, $g$ averages $1.23$ and ranges from $0.77$ to $1.74$. On the $15$ test scenes, it is close to $1$. With the nominal ratio, the training targets are thus on average $1.23$ times brighter than the amplified inputs. A network trained on these pairs learns this brightening and also applies it to the test scenes. Calibration amplifies each training scene by $50g$ instead of $50$; the test data are unchanged. It lifts every network by $3.2$ to $5.8$\,dB (Tab.~\ref{tab:hsidwrd_calib}). \ours{} keeps the best MPSNR and SSIM, $0.19$\,dB above HSDT, $0.75$\,dB above SST and $0.87$\,dB above SERT; SSUMamba has the best SAM.

\textbf{HIR-Diff on real noise.} HIR-Diff is not trained on HSIDwRD, so its output follows the exposure of the test input and carries no brightening learned from the training scenes in Tab.~\ref{tab:hsidwrd}. This is consistent with its nominal MPSNR, $0.02$\,dB above \ours{} and $1.6$ to $3.3$\,dB above the other retrained networks, while its SAM ($4.663$) is more than twice of \ours{} and its SSIM is below \ours{} and SST. Once the training pairs are calibrated, all five networks of Tab.~\ref{tab:hsidwrd_calib} exceed it by $2.3$ to $3.2$\,dB.
\begin{table}[t]
\centering
\caption{Exposure calibration of the HSIDwRD training pairs: each model
retrained with the nominal $\times50$ exposure ratio and with the ratio
fitted per training scene; test data unchanged.}
\label{tab:hsidwrd_calib}
\footnotesize
\setlength{\tabcolsep}{3pt}
\resizebox{\linewidth}{!}{%
\begin{tabular}{@{}l ccc ccc r@{}}
\toprule
& \multicolumn{3}{c}{Nominal} & \multicolumn{3}{c}{Calibrated} & \\
\cmidrule(lr){2-4}\cmidrule(lr){5-7}
Method & MPSNR$\uparrow$ & SSIM$\uparrow$ & SAM$\downarrow$ & MPSNR$\uparrow$ & SSIM$\uparrow$ & SAM$\downarrow$ & $\Delta$ \\
\midrule
SERT~\cite{sert} & 27.716 & 0.9267 & 2.840 & 32.424 & 0.9670 & 2.545 & $+4.71$ \\ 
SST~\cite{sst} & 28.416 & 0.9490 & 2.821 & 32.543 & 0.9667 & 2.523 & $+4.13$ \\ 
SSUMamba~\cite{ssumamba} & 26.799 & 0.8831 & 2.539 & 32.552 & 0.9688 & \textbf{2.075} & $+5.75$ \\ 
HSDT~\cite{hsdt} & 28.508 & 0.9465 & 2.740 & 33.096 & 0.9699 & 2.302 & $+4.59$ \\ 
\textbf{\ours{}} & \textbf{30.099} & \textbf{0.9606} & \textbf{2.181} & \textbf{33.289} & \textbf{0.9705} & 2.089 & $+3.19$ \\ 
\bottomrule
\end{tabular}}
\end{table} 
\begin{table*}[t]
\centering
\caption{Class ablation: all static ($\mathsf{B}$, $\mathsf{K}$), all
gated ($\mathsf{S}$, $\mathsf{G}$) or all adaptive ($\mathsf{R}$,
$\mathsf{A}$) components of the backbone are replaced by the identity and the network is retrained using the full-model recipe; last iterate.}
\label{tab:class_ablate}
\footnotesize
\setlength{\tabcolsep}{4pt}
\begin{tabular}{@{}l rccc rccc rccc@{}}
\toprule
& \multicolumn{4}{c}{ICVL, $\sigma{=}50$} & \multicolumn{4}{c}{HSIDwRD} & \multicolumn{4}{c}{Pavia Centre $\times4$} \\
\cmidrule(lr){2-5}\cmidrule(lr){6-9}\cmidrule(lr){10-13}
Arm & Params & MPSNR$\uparrow$ & SSIM$\uparrow$ & SAM$\downarrow$ & Params & MPSNR$\uparrow$ & SSIM$\uparrow$ & SAM$\downarrow$ & Params & MPSNR$\uparrow$ & SSIM$\uparrow$ & SAM$\downarrow$ \\
\midrule
Full \ours{} & 5.450M & \textbf{42.927} & \textbf{0.9734} & 1.304 & 5.450M & \textbf{30.099} & \textbf{0.9606} & \textbf{2.181} & 6.505M & 29.632 & 0.8410 & 5.111 \\ 
Without static & 4.250M & 29.406 & 0.6994 & 15.526 & 4.251M & 29.501 & 0.9500 & 2.263 & 3.156M & \textbf{29.814} & \textbf{0.8477} & \textbf{5.023} \\ 
Without gated & 3.825M & 42.618 & 0.9719 & \textbf{1.290} & 3.825M & 28.131 & 0.9348 & 2.362 & 4.973M & 28.967 & 0.8114 & 5.561 \\ 
Without adaptive & 2.830M & 34.285 & 0.8893 & 5.336 & 2.831M & 28.063 & 0.9482 & 2.565 & 5.225M & 29.600 & 0.8392 & 5.164 \\ 
\bottomrule
\end{tabular}
\end{table*}
\begin{table}[t]
\centering
\caption{Component ablation of the ICVL denoiser: MPSNR on the 50 protocol crops, $\Delta$ is at $\sigma=50$.}
\label{tab:icvl_ablate}
\footnotesize
\setlength{\tabcolsep}{3pt}
\resizebox{\linewidth}{!}{%
\begin{tabular}{@{}l r ccccc r@{}}
\toprule
& & \multicolumn{5}{c}{MPSNR$\uparrow$} & \\
\cmidrule(lr){3-7}
Arm & Params & $\sigma{=}10$ & $\sigma{=}30$ & $\sigma{=}50$ & $\sigma{=}70$ & Blind & $\Delta$ \\
\midrule
Full \ours{} ($N{=}2$, $k{=}24$) & 5.450M & 49.209 & 45.131 & 42.927 & 41.326 & 43.263 & -- \\ 
Dense MLP for tensorial summation$^{a}$ & 5.450M & 48.871 & 45.065 & 42.846 & 41.166 & 43.179 & $-0.08$ \\ 
No window shift in SW-MSA & 5.450M & 49.091 & 44.849 & 42.626 & 40.996 & 42.930 & $-0.30$ \\ 
No strip-window coupling $\mathsf{K}$ & 5.338M & 47.718 & 41.197 & 35.377 & 31.740 & 36.039 & $-7.55$ \\ 
No scan $\mathsf{S}$ & 5.244M & 49.104 & 45.117 & 42.883 & 41.280 & 43.187 & $-0.04$ \\ 
No pooled spectral attention $\mathsf{R}$ & 5.376M & 49.244 & 45.191 & 42.965 & 41.341 & 43.278 & $+0.04$ \\ 
$N{=}6$, $k{=}1$$^{b}$ & 4.939M & 47.863 & 44.288 & 42.250 & 40.770 & 42.592 & $-0.68$ \\ 
\bottomrule
\end{tabular}}
\begin{flushleft}\footnotesize
$^{a}$ Iso-parameter residual MLP in the tensorial-summation layers.
$^{b}$ Six blocks with one attention unit each ($k=1$ instead of $24$): $6$ attention units in total instead of $48$, and $12$ tensorial-summation layers instead of $4$.
\end{flushleft}
\end{table}
\begin{table}[t]
\centering
\caption{Component ablation on HSIDwRD (nominal training pairs), $\Delta$ is in MPSNR.}
\label{tab:hsidwrd_ablate}
\footnotesize
\setlength{\tabcolsep}{3pt}
\resizebox{\linewidth}{!}{%
\begin{tabular}{@{}l r ccc r@{}}
\toprule
Arm & Params & MPSNR$\uparrow$ & SSIM$\uparrow$ & SAM$\downarrow$ & $\Delta$ \\
\midrule
Full \ours{} ($N{=}2$, $k{=}24$) & 5.450M & \textbf{30.099} & \textbf{0.9606} & 2.181 & -- \\ 
Dense MLP for tensorial summation & 5.451M & 28.142 & 0.9474 & 2.214 & $-1.96$ \\ 
No window shift in SW-MSA & 5.450M & 29.166 & 0.9553 & \textbf{2.167} & $-0.93$ \\ 
No strip-window coupling $\mathsf{K}$ & 5.338M & 29.628 & 0.9597 & 2.211 & $-0.47$ \\ 
No scan $\mathsf{S}$ & 5.245M & 28.378 & 0.9520 & 2.397 & $-1.72$ \\ 
No pooled spectral attention $\mathsf{R}$ & 5.376M & 28.878 & 0.9566 & 2.226 & $-1.22$ \\ 
$N{=}6$, $k{=}1$ & 4.940M & 28.412 & 0.9527 & 2.486 & $-1.69$ \\ 
\bottomrule
\end{tabular}}
\end{table}
\begin{table}[t]
\centering
\caption{Component ablation on Pavia Centre $\times4$: last iterate of
the 100k schedule; $\Delta$ in MPSNR.}
\label{tab:pavia_ablate}
\footnotesize
\setlength{\tabcolsep}{3pt}
\resizebox{\linewidth}{!}{%
\begin{tabular}{@{}l r cccccc r@{}}
\toprule
Arm & Params & MPSNR$\uparrow$ & SSIM$\uparrow$ & SAM$\downarrow$ & CC$\uparrow$ & ERGAS$\downarrow$ & RMSE$\downarrow$ & $\Delta$ \\
\midrule
Full \ours{} ($N{=}8$, $k{=}3$) & 6.505M & 29.632 & 0.8410 & 5.111 & 0.9105 & 5.491 & 0.0341 & -- \\ 
No window shift in SW-MSA & 6.505M & 29.545 & 0.8388 & 5.129 & 0.9089 & 5.542 & 0.0345 & $-0.09$ \\ 
$N{=}8$, $k{=}1$ & 5.376M & \textbf{29.722} & \textbf{0.8448} & \textbf{5.100} & \textbf{0.9124} & \textbf{5.433} & \textbf{0.0337} & $+0.09$ \\ 
\bottomrule
\end{tabular}}
\end{table}
\begin{figure*}[t]
\centering
\includegraphics[width=\textwidth]{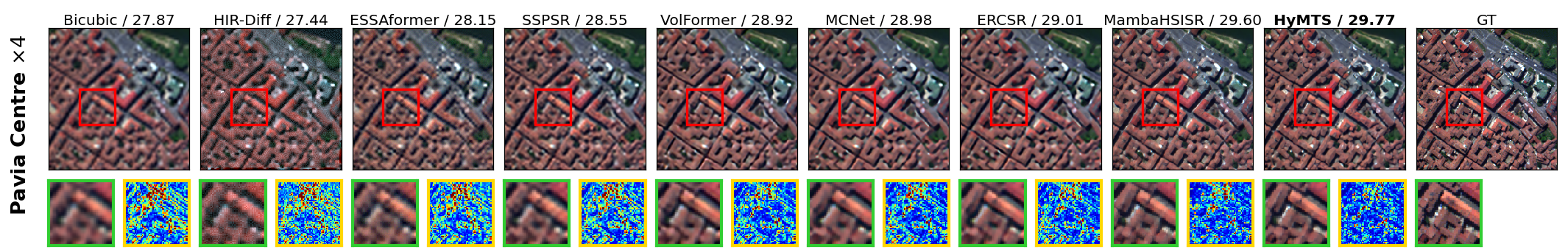}\\[2pt]
\label{fig:qual_pavia}
\includegraphics[width=\textwidth]{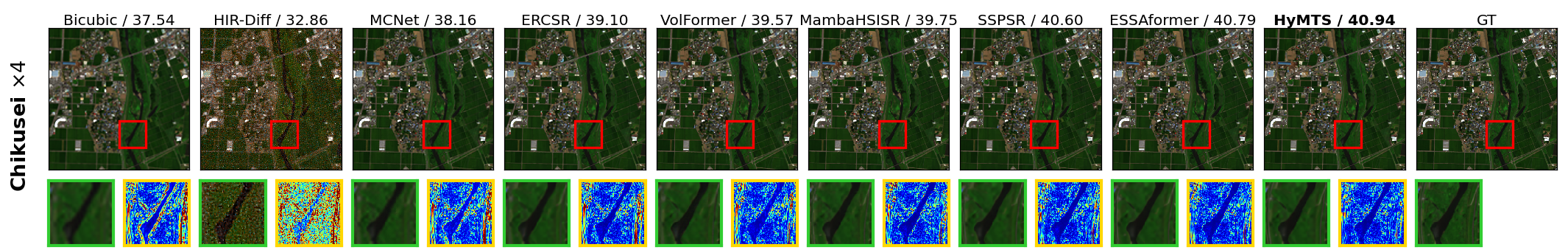}
\caption{$\times4$ super-resolution on a Pavia Center test image (bands 52, 28, 7) and on a Chikusei test image (bands 56, 36, 17). In every image, the red box marks a region that is shown enlarged in the green box below it; the yellow box next to it is the absolute error of the enlarged region, averaged over the bands (dark blue is zero, dark red is $0.08$ or more on Pavia Center and $0.03$ or more on Chikusei).}
\label{fig:qual_sr}
\end{figure*}
\section{Complexity measurement}
\label{sec:supp_cost}
\noindent\textbf{Reference size.} All values refer to one $512\times512$ output cube. For denoising, the input is one $512\times512$ noisy cube with the dataset's band count (31 on ICVL, 34 on HSIDwRD). For super-resolution, the input is one $32\times32$ low-resolution cube with the band count of the dataset (31 on CAVE, 128 on Chikusei, 102 on Pavia Center), and the count is multiplied by $(512/(32s))^2$ for the scale factor
$s$.

\noindent\textbf{Exceptions.} The selective-scan and causal-convolution kernels of SSUMamba and MambaHSISR are compiled extensions that the counter does not see; we add them analytically, $6\,b\,d\,l\,n$ MACs per scan over a sequence of length $l$ with batch $b$, width $d$ and state size $n$, and $4\,b\,d\,l$ per causal convolution. SSUMamba is run through its own
inference on $64\times64$ blocks, as in the tables. For VolFormer in
super-resolution, we count the spectral path that is executed, without its
RGB decoder and the layers that the forward pass does not reach. HIR-Diff
samples a three-channel reduced image at the output resolution, so its
cost is that of its 20 sampling steps, one evaluation of the pretrained
diffusion network each, and does not depend on the band count.
\section{Additional visual results}
\label{sec:supp_visual}
Fig.~\ref{fig:qual_sr} shows one test image of Pavia Center at $\times4$ (top) and Chikusei at $\times4$ (bottom) for each method in the corresponding table. On Pavia Center, the roof edges of the enlarged region are sharpest for MambaHSISR and \ours{}, and \ours{} has the lower error ($29.77$ against $29.60$\,dB on the image). On Chikusei, all networks' errors lie on the boundary of the dark channel and on the field borders. \ours{} is $0.15$\,dB
above ESSAformer on the image.

\end{document}